%% file: 0_main.tex
\documentclass[10pt,twocolumn,letterpaper]{article}

\usepackage[pagenumbers]{cvpr} 

\usepackage[accsupp]{axessibility}  
\usepackage{graphicx}
\usepackage{amsmath}
\usepackage{amssymb}
\usepackage{booktabs}
\usepackage{tabularx}
\usepackage{float}
\usepackage{multirow}
\usepackage{siunitx}
\usepackage{algorithm}
\usepackage{algorithmic}
\usepackage{makecell} 


\definecolor{cvprblue}{rgb}{0.21,0.49,0.74}
\usepackage[pagebackref,breaklinks,colorlinks,allcolors=cvprblue]{hyperref}

\def\paperID{8076} 
\def\confName{CVPR}
\def\confYear{2025}

\newcommand{\li}{Li \emph{et al.}~\cite{cis2020}} 
\newcommand{\zhu}{Zhu \emph{et al.}~\cite{zhu2022montecarlo}} 
\newcommand{\garon}{Garon \emph{et al.}~\cite{garon19}}

\title{Channel-wise Noise Scheduled Diffusion for Inverse Rendering in Indoor Scenes}

\author{
{JunYong Choi$^{1,2}$}\quad 
{Min-cheol Sagong$^{1}$}\quad 
{SeokYeong Lee$^{1,2}$}\quad
{Seung-Won Jung$^{2}$}\quad\\
{Ig-Jae Kim$^{1,3,4}$}\quad 
{Junghyun Cho$^{1,3,4}$} 
\\[2mm]
{$^{1}$Korea Institute of Science and Technology(KIST)}\quad 
{$^{2}$Korea University} \\
{$^{3}$AI-Robotics, KIST School, University of Science and Technology} \\
{$^{4}$Yonsei-KIST Convergence Research Institute} \\
\vspace{-3mm}
{\tt\small \{happily,mcsagong,shapin94,drjay,jhcho\}@kist.re.kr}\quad
{\tt\small swjung83@korea.ac.kr}
}

\begin{document}

\input{1_abstract}
\input{2_intro_related}

\input{3_method}
\input{4_experiments}

\input{5_supplementary}

{
\small
\bibliographystyle{ieeenat_fullname}
\bibliography{egbib}
}

\end{document}

%% file: 1_abstract.tex
\twocolumn[{
\renewcommand\twocolumn[1][]{#1}%
\maketitle
\begin{center}
    \centering
    \captionsetup{type=figure}
    \includegraphics[width=\textwidth]{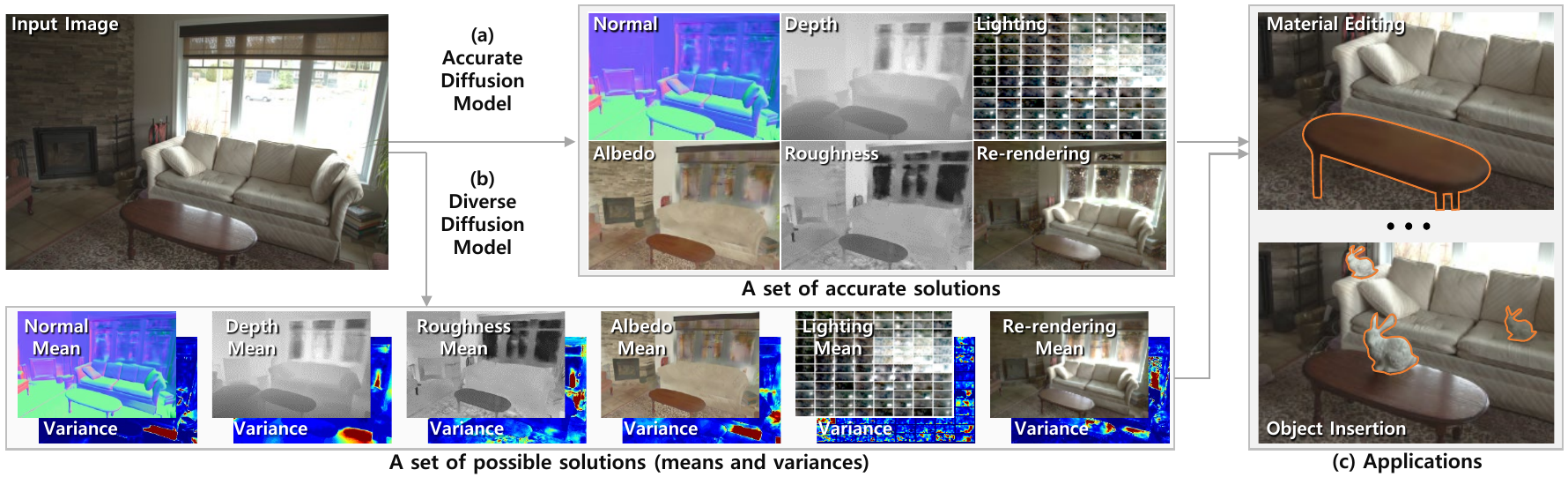}
    \vspace{-15pt}
    \captionof{figure}{\textbf{Diffusion-Based Inverse Rendering.} 
    We present a diffusion-based inverse rendering framework that addresses the two competing goals of accuracy and diversity in inverse rendering by using two distinct models, each dedicated to achieving one of the objectives.
    (a) Our first model predicts a single accurate solution, while (b) the second model presents diverse possible solutions. (c) This enables practical applications such as object insertion and material editing (e.g., increasing the roughness of the brown table).}
    \label{fig:main} 
\end{center}
}]

\begin{abstract}
We propose a diffusion-based inverse rendering framework that decomposes a single RGB image into geometry, material, and lighting. Inverse rendering is inherently ill-posed, making it difficult to predict a single accurate solution. To address this challenge, recent generative model-based methods aim to present a range of possible solutions. However, finding a single accurate solution and generating diverse solutions can be conflicting. In this paper, we propose a channel-wise noise scheduling approach that allows a single diffusion model architecture to achieve two conflicting objectives. The resulting two diffusion models, trained with different channel-wise noise schedules, can predict a single highly accurate solution and present multiple possible solutions. The experimental results demonstrate the superiority of our two models in terms of both diversity and accuracy, which translates to enhanced performance in downstream applications such as object insertion and material editing. 
\end{abstract}

%% file: 2_intro_related.tex
\section{Introduction}
Single-image inverse rendering is a long-standing challenge in computer vision and computer graphics that involves the decomposition of an RGB image into its constituent elements of material, lighting, and geometry. This technology enables a wide range of applications, including object insertion, material editing, and relighting. Although inverse rendering has a clear solution, interpreting a single observed radiance can be complex, as it may result from various combinations of material, geometry, and lighting. For instance, a high radiance might be due to the material being white, intense lighting, or a pronounced specular component caused by the Fresnel effect. This inherent ill-posedness of inverse rendering introduces ambiguity and difficulty, particularly when working with a single-image. Therefore, it is important not only to predict an optimal solution, but also to propose multiple potential solutions. However, as discussed in other studies~\cite{lee2019harmonizing, yu2019free}, enforcing only the reconstruction loss cannot improve the sample diversity, leading to conflicts between the objectives.

Learning-based methods~\cite{cis2020, irisformer2022, zhu2022montecarlo, choi2023mair, mair++} leveraging large-scale synthetic datasets have demonstrated excellent performance. Strong supervised signals from reconstruction losses on well-aligned ground truths have enabled these methods to effectively decompose scenes. However, because these methods are based on deterministic approaches, they do not account for the solution space of possible combinations. Recently, inverse rendering studies based on Latent Diffusion Models (LDM)~\cite{ldm, podell2023sdxl} have emerged~\cite{luo2024intrinsicdiffusion, zeng2024rgb, kocsis2023intrinsic}. These probabilistic methods, which rely on priors learned from large image-text datasets, not only provide multiple solutions but also exhibit excellent predictive capability. However, they often overlook the trade-off between sample diversity and prediction accuracy and they struggle to account for the dependencies between modalities. Additionally, they rely on the pretrained three-channel autoencoder of LDM, which is effective at generating image-based data such as RGB images but falls short in handling high-dimensional data like per-pixel lighting, which is crucial for applications such as photorealistic object insertion and material editing.

While probabilistic methods are effective for capturing multiple potential solutions in ill-posed problems like inverse rendering, they can be sub-optimal for predicting a definitive ground truth. We found that diffusion models exhibit a trade-off between diversity and accuracy depending on the entire timestep \( T \). Motivated by these observations, we propose two conditional diffusion models, each focusing on either diversity or accuracy. Our diffusion models predict multiple modalities, such as a depth map, normal map, diffuse albedo map, roughness map, and per-pixel environment map, simultaneously from a single RGB image. The first proposed model, the \textbf{P}rogressive \textbf{D}iffusion \textbf{M}odel (PDM), focuses on diversity using a large \( T \) for training. The second model, the \textbf{S}witchable \textbf{D}iffusion \textbf{M}odel (SDM), is trained with a smaller \( T \) to prioritize accuracy. To improve the quality of inverse rendering, a channel-wise noise scheduling approach is introduced to adjust the transition between modalities during generation.

Our PDM and SDM operate in a low-resolution pixel space rather than a latent space, allowing them to handle per-pixel environment maps. To mitigate the high computational costs associated with handling high-dimensional data directly, we introduce a method for encoding these maps into compact feature vectors. Additionally, we present a simple RGB-guided super-resolution model for up-sampling outputs.

We evaluated the performance by comparing it with other inverse rendering baselines on synthetic~\cite{choi2023mair} and real-world~\cite{garon19, wu2023maw, vasiljevic2019diode} datasets. The experimental results showed that SDM outperforms in images with fewer ambiguous regions, where accurate predictions are more feasible, while PDM excels in complex scenes with numerous ambiguous regions requiring diverse predictions, indicating the complementary nature of SDM and PDM. In summary, our main contributions are as follows:

\begin{enumerate}
    \item We propose a novel diffusion-based inverse rendering framework that incorporates per-pixel environment maps while considering the correlations between modalities.
    \item We propose channel-wise noise scheduling, which improves generation quality by adjusting the modality generation transition in multi-modality diffusion models.
    \item We demonstrate superior performance over existing baselines on both real and synthetic datasets, highlighting the importance of considering both diversity and accuracy in inverse rendering.   
\end{enumerate}

\begin{figure*}[ht]
  \centering
  \includegraphics[width=\linewidth]{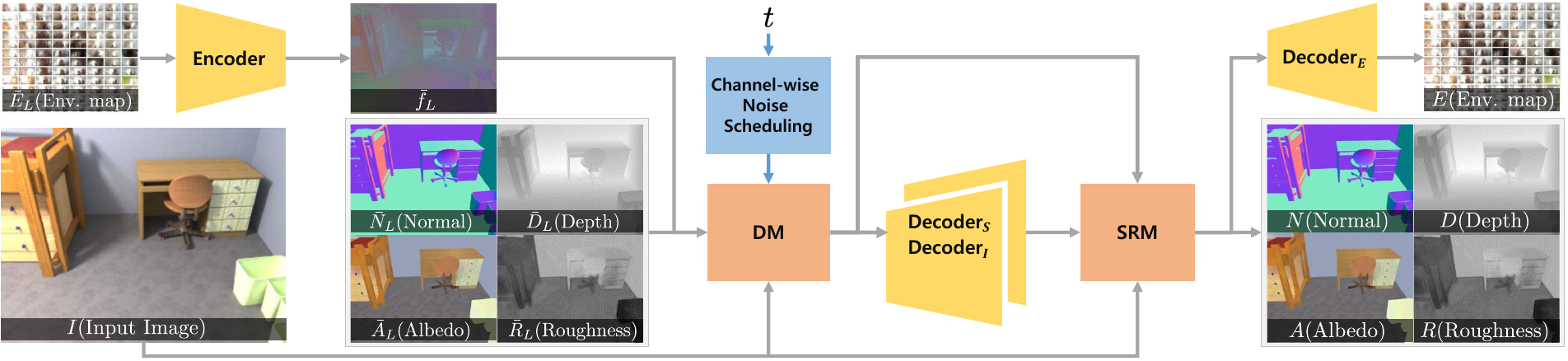}
  \caption{\textbf{Entire pipeline.} Our channel-wise noise scheduling assists the DM's inference by adjusting the transitions between modalities based on the timestep. Since the DM operates at a low resolution, modalities are up-sampled through our SRM.}
  \label{fig:overview}
\end{figure*}

\section{Related Works}
\noindent\textbf{Inverse rendering.}
Inverse rendering involves the extraction of depth, normal, SVBRDF, and lighting information from RGB images. Initial research concentrated primarily on specific areas, such as shape-from-shading~\cite{sfs1,sfs2,sfs3} and intrinsic image decomposition~\cite{id1,id2,iiw,luo2020niid}. Advances in deep learning have facilitated the study of complex properties, such as SVBRDF~\cite{sime1, sime2, li2018learning, sime4} and lighting~\cite{gardner2019deep, garon19, vsg, lighthouse}. Li \emph{et al.}~\cite{cis2020, openrooms2021} introduced OpenRooms and a spherical-Gaussian lighting model. IRISformer~\cite{irisformer2022} leverages long-range attention through a dense vision transformer. \zhu\ introduced the refined InteriorVerse dataset and incorporated Monte Carlo ray tracing into the inverse rendering. Choi \emph{et al.}~\cite{choi2023mair, mair++} extended this approach to a multi-view setup. However, all of these methods rely on deterministic models, limiting their ability to propose alternative solutions. Our work diverges by proposing PDM and SDM, capable of generating multiple solutions and surpassing deterministic models in flexibility and accuracy.

\noindent\textbf{Diffusion models.}
Diffusion models~\cite{ddpm, adm, song2019generative}, which have recently emerged as powerful tools for image generation, are now being applied to various tasks. Several studies have trained diffusion models directly on target datasets to generate materials~\cite{vecchio2024matfuse}, distant lighting~\cite{lyu2023diffusion, phongthawee2024diffusionlight}, and human face reflectance~\cite{papantoniou2023relightify}. Recently, research has begun exploring inverse rendering by utilizing the strong priors of pretrained LDMs. Kocsis \emph{et al.}~\cite{kocsis2023intrinsic} fine-tuned an LDM on synthetic datasets, enabling the generation of multiple plausible material maps. RGB$\leftrightarrow$X~\cite{zeng2024rgb} demonstrated that LDMs can be used for both inverse rendering and rendering. Luo \emph{et al.}~\cite{luo2024intrinsicdiffusion} trained ControlNet~\cite{zhang2023adding} to predict multiple intrinsic modalities. However, since these methods predict each modality independently, they fail to address inter-modality dependencies and struggle with environment maps, limiting realistic lighting inference and applications. In contrast, our framework accounts for these dependencies, enhancing sample fidelity, while per-pixel lighting enables realistic, high-fidelity object insertion.



%% file: 3_method.tex
\section{Method}
Given an input RGB image, denoted as \(\mathrm{I} \in \mathbb{R}^{3 \times H \times W}\), our goal is to obtain the following outputs: a normal map \(\mathrm{N} \in \mathbb{R}^{3 \times H \times W}\), a depth map \(\mathrm{D} \in \mathbb{R}^{H \times W}\), a diffuse albedo map \(\mathrm{A} \in \mathbb{R}^{3 \times H \times W}\), a roughness map \(\mathrm{R} \in \mathbb{R}^{H \times W}\), and a per-pixel environment map, represented as \(\mathrm{E} \in \mathbb{R}^{3 \times C \times H \times W}\), where \( H \) and \( W \) represent the image dimensions, and \( C \) denotes the angular resolution of the environment map. First, we introduce the synthetic dataset used in our experiments. Second, we introduce a method to encode per-pixel lighting into a neural feature vector. Next, we introduce the conditional diffusion model that serves as the foundation of our approach, along with two channel-wise noise scheduling techniques that allow us to customize the model for our desired application. Since the diffusion model operates in the low-resolution pixel space, we introduce a RGB-guided super-resolution model for upsampling. The entire pipeline is shown in Fig.~\ref{fig:overview}.

\subsection{Dataset}
To train our conditional diffusion model, we need well-aligned ground truth scene information (\(\mathrm{N}, \mathrm{D}, \mathrm{A}, \mathrm{R}, \mathrm{E}\)) pairs for RGB images. For this purpose, two synthetic datasets, OpenRooms~\cite{openrooms2021} and InteriorVerse~\cite{zhu2022montecarlo}, are suitable; however, the lighting part of InteriorVerse is currently not available. Additionally, OpenRooms often contains overly simplistic images (such as views of walls or incorrectly placed light sources), which hinder the diffusion model from learning the data distribution. Therefore, we used the OpenRooms FF dataset~\cite{choi2023mair}, which involves filtering out inappropriate images from OpenRooms.

\subsection{Implicit Lighting Representation}
As a method for handling high-dimensional data such as per-pixel environment maps, a recent method called MAIR++~\cite{mair++} introduced the implicit lighting representation (ILR), which represents environment maps as neural feature vectors. MAIR++ processes the RGB, geometry, and material using a CNN to obtain the ILR, which is subsequently processed by three different decoders to obtain an environment map, shading map, and specular radiance. This approach can render images in a realistic and computationally efficient way. We are motivated to use the ILR as our lighting representation. However, since the previous ILR~\cite{mair++} is not inferred from the environment map, it neither accurately reproduces the ground truth lighting nor effectively captures the detailed high-dynamic range (HDR) radiance values. In contrast, we re-design the ILR as an MLP-based autoencoder that takes the environment map as input. 

Since the HDR radiance values of the environment map cover a very wide range, using them directly as input causes the network to diverge easily. Therefore, we transform the input into the log space using \texttt{log1p}~\cite{press2007numerical} and add batch normalization to the encoder MLP. These two simple techniques allow the network to handle HDR lighting more effectively, leading to significantly improved performance. Mathematically, the encoder is defined as follows:

\begin{equation}\label{eqn:eq_encoder}
\boldsymbol{f} = \text{tanh}(\text{Encoder}(\text{log1p}(\bar{\mathrm{E}}))), 
\end{equation}
where \(\bar{\cdot}\) represents the ground truth, and \(\boldsymbol{f} \in \mathbb{R}^{96}\) is the ILR neural feature vector. To enable realistic rendering through \(\boldsymbol{f}\), the three decoders are applied:

\vspace{-2mm}
\begin{equation}\label{eqn:eq_decoders}
\begin{aligned}
    \mathrm{E} &= \text{expm1}(\text{Decoder}_E(\boldsymbol{f})), \\
    \mathrm{S} &= \text{expm1}(\text{Decoder}_S(\boldsymbol{f})), \\
    \mathrm{I}_s &= \text{expm1}(\text{Decoder}_I(\boldsymbol{f}, \bar{\mathrm{R}}, \gamma(\boldsymbol{r}), \bar{\mathrm{N}} \cdot \boldsymbol{v})).
\end{aligned}
\end{equation}
where \texttt{expm1}~\cite{press2007numerical} is the inverse of \texttt{log1p}, \(\mathrm{S} \in \mathbb{R}^{3 \times H \times W}\) denotes the shading map, \(\mathrm{I}_s \in \mathbb{R}^{3 \times H \times W}\) represents the specular image, \(\gamma\) refers to the frequency positional encoding, and \(\boldsymbol{r}\) and \(\boldsymbol{v}\) indicate the reflection direction and the viewing direction, respectively. Through these three decoders, \(\boldsymbol{f}\) can not only reconstruct the environment map but also be used to render arbitrary views. 

\begin{table*}[ht]
\centering
\resizebox{\linewidth}{!}{
\begin{footnotesize} 
\begin{tabular}{|c||c c|c c|c c|c c|c c|c c|}
\hline
\multirow{2}{*}{\(T\)} & \multicolumn{2}{c|}{\(\mathrm{N}_L\)} & \multicolumn{2}{c|}{\(\mathrm{D}_L\)} & \multicolumn{2}{c|}{\(\mathrm{A}_L\)} & \multicolumn{2}{c|}{\(\mathrm{R}_L\)} & \multicolumn{2}{c|}{\(\mathrm{E}_L\)} & \multicolumn{2}{c|}{\(\mathrm{I}_L\)} \\ \cline{2-13}
                       & \(\text{MSE}\downarrow\) & \(\text{var}\uparrow \) & \(\text{MSE}\downarrow\) & \(\text{var}\uparrow \) & \(\text{MSE}\downarrow\) & \(\text{var}\uparrow \) & \(\text{MSE}\downarrow\) & \(\text{var}\uparrow \) & \(\text{MSE}\downarrow\) & \(\text{var}\uparrow \) & \(\text{MSE}\downarrow\) & \(\text{var}\uparrow \) \\ \hline
1000   & 2.236  & \textbf{0.183} & \underline{0.065}  & \textbf{0.020}  & \underline{0.418} & \textbf{0.034} & 5.518  & \textbf{0.106} & \underline{12.66} & \textbf{79.61} & \underline{0.179} & \textbf{0.042} \\ \hline
100    & \underline{2.238} & 0.182 & 0.064  & 0.019 & 0.413 & \textbf{0.034} & \textbf{5.354}  & 0.094 & \underline{12.66} & 77.62 & 0.177 & 0.041 \\ \hline
10     & 2.026  & 0.138 & \textbf{0.059}  & 0.013 & 0.374 & 0.024 & 5.440  & 0.066 & 12.13 & 62.38 & 0.162 & 0.03 \\ \hline
1      & \textbf{1.817}  & \underline{0.0}  & 0.061  & \underline{0.0}   & \textbf{0.334} & \underline{0.0}  & \underline{5.545}  & \underline{0.0}  & \textbf{11.37} & \underline{0.06} & \textbf{0.137} & \underline{0.0} \\ \hline
\end{tabular}
\end{footnotesize} 
}
\caption{\textbf{Ablation of training \( T \).} The best values in each column are highlighted in \textbf{bold}, and the worst values are \underline{underlined}. In most cases, as the value of \( T \) decreases, prediction accuracy improves, but the diversity of the samples is reduced. To evaluate under general conditions, we fixed \( \tau \) to 1.0 for all settings, and the DDIM~\cite{ddim} steps were selected as 2, 2, 2, and 1, respectively.}
\label{tab:T_exp}
\end{table*}

\subsection{Diffusion-based Inverse Rendering}
Since the introduction of LDM~\cite{ldm, podell2023sdxl}, many diffusion-based methods~\cite{luo2024intrinsicdiffusion, zeng2024rgb, kocsis2023intrinsic} have used pretrained LDMs that have strong priors for images. However, these methods do not provide meaningful priors for high-dimensional data such as environment maps. Therefore, we adopt a cascaded structure~\cite{ho2022cascaded} that first generates $\mathbf{z}_0=(\mathrm{N}_L, \mathrm{D}_L, \mathrm{A}_L, \mathrm{R}_L, \boldsymbol{f}_L)$ and then increases the resolution, where the subscript $L$ denotes a low-resolution image of size \(H_L \times W_L\). During inference, our conditional diffusion model (DM) takes \( \mathrm{I}_L \) as a condition and iteratively denoises the Gaussian noise image \( \mathbf{z}_{T-1} \sim \mathcal{N}(0, 1) \) to produce \( \mathbf{z}_0 \). During training, we predict the velocity \( \mathbf{v}_t \)~\cite{salimans2022progressive}, and the loss function $L_{\text{DM}}$ is given by:

\begin{equation}
    L_\text{DM} = \lVert \mathbf{v}_t - \text{DM}(t, \mathbf{z}_t, \mathrm{I}_L, \text{CLIP}(\mathrm{I}_L)) \rVert_2^2, 
\end{equation} 
\vspace{-2mm}
\begin{equation}
    \mathbf{v}_t = \sqrt{\bar{\alpha}_t} \mathbf{z}_{T-1} - \sqrt{1 - \bar{\alpha}_t} \mathbf{z}_0,
\end{equation}
where \( T \) is the entire timestep, \( t \in [0, T-1] \) is uniformly sampled during training, \( \bar{\alpha}_t \) is the noise scheduler, \( \mathbf{z}_t = \sqrt{\bar{\alpha}_t} \mathbf{z}_0 + \sqrt{1 - \bar{\alpha}_t} \mathbf{z}_{T-1} \), and \(\text{CLIP}\) is the pretrained image encoder~\cite{clip}. \( \mathrm{I}_L \) is concatenated with \( \mathbf{z}_t \), and the CLIP embedding is used as a cross-attention condition.

\begin{figure}[ht]
  \centering
  \includegraphics[width=\linewidth]{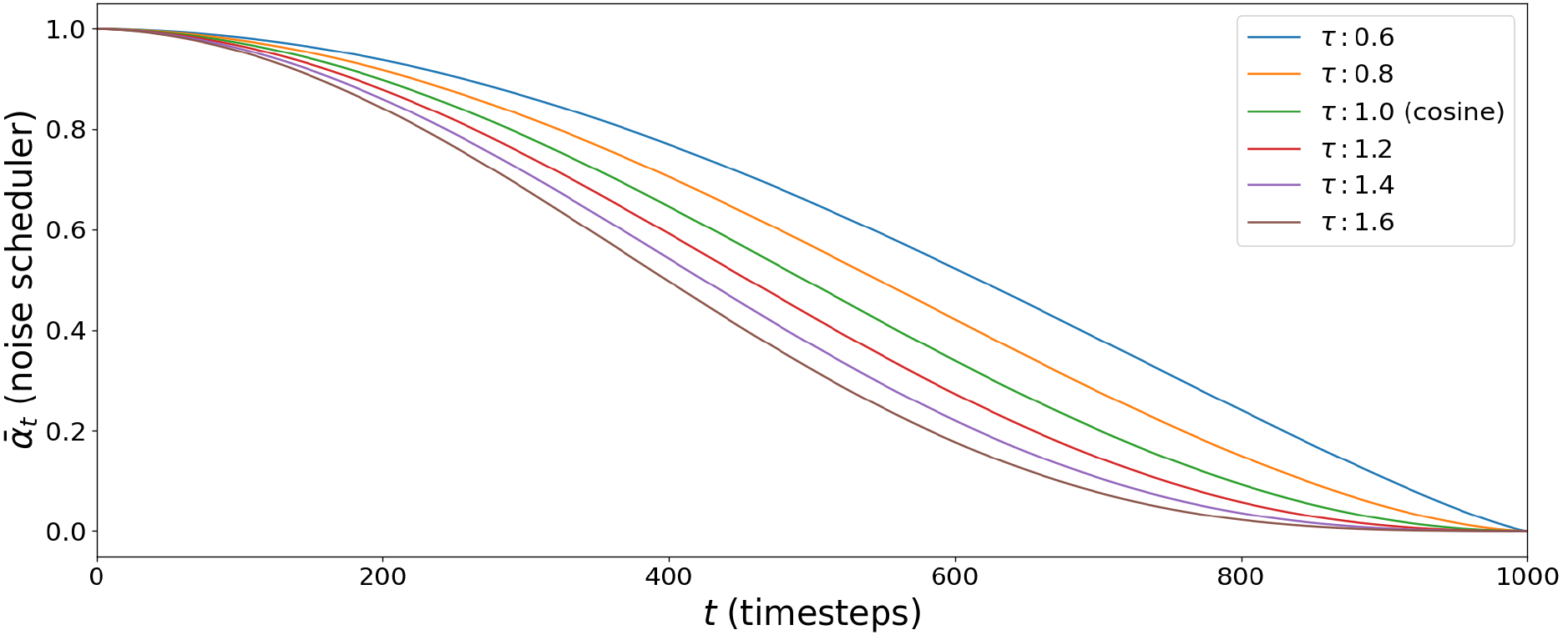}
  \caption{\textbf{Parameterizing noise scheduler.} A larger \( \tau \) value results in a slower generation of the modality.}
  \label{fig:tau_noise}
\end{figure}

\noindent{\bf Channel-wise noise scheduling.} 
Since we generate multiple interdependent modalities simultaneously, we hypothesize that there is an optimal transition to generate these modalities in inverse rendering. For example, according to previous works~\cite{zhang2021nerfactor, li2018learning, yu2019inverserendernet}, geometry typically has the least uncertainty and supports the prediction of material and lighting, so generating geometry first could be advantageous. On the other hand, since lighting has the highest uncertainty, generating it last could be beneficial. To validate this hypothesis, we divide \( \mathbf{z}_0 \) into three categories based on the modality attributes: geometry (\(\mathrm{N}_L, \mathrm{D}_L\)), material (\(\mathrm{A}_L, \mathrm{R}_L\)), and lighting (\(\boldsymbol{f}_L\)), and accordingly apply different noise scheduling to the three parts. Specifically, following Chen \emph{et al.}~\cite{chen2023importance}, we select the cosine scheduler as the base noise scheduler and parameterize it as follows:

\vspace{-2mm}
\begin{equation}
\begin{split}
    v_s &= \cos(s \cdot 2 \pi)^{2\tau}, \quad v_b = \cos(b \cdot 2 \pi)^{2\tau} \\
    \alpha_t &= \cos(((b-s)\cdot t + s) \cdot 2 \pi)^{2\tau}, \bar{\alpha}_t = \frac{v_b - \alpha_t}{v_b - v_s}.
\end{split}
\end{equation}
Throughout the paper, we fix \( s = 0.008 \) and \( b = 1 \) and control the scheduler using only \( \tau \). When \( \tau = 1 \), $\bar{\alpha}_t$ is equivalent to the cosine scheduler. The behavior of the noise scheduler as a function of \( \tau \) is shown in Fig.~\ref{fig:tau_noise}.

Fig.~\ref{fig:tau_noise} shows that as the value of \( \tau \) increases, the noise increases more gradually. To determine the optimal modality generation order using $\tau$, we conducted experiments by applying different values of \( \tau \) to geometry, material, and lighting, meaning that different noise levels are applied separately to each channel of the data during the denoising process of the diffusion model. The ablation study in Tab.~\ref{tab:tau_experiments} confirms that generating in the order of geometry, material, and lighting is the most effective, aligning with our intuition. As a result, through our channel-wise noise scheduling, it focuses on specific modalities at different timesteps, enabling more effective predictions of other modalities in subsequent timesteps.

In general, \( T \) in diffusion models is set to a large number, such as 1000. This helps approximate the data distribution and increases the diversity of the generated data~\cite{iddpm}. However, we note that such diffusion models are challenging in predicting a single definitive ground truth. To this end, we conducted an ablation study on \( T \) values by training the DM with \( T = 1000, 100, 10, \text{ and } 1 \). For each setting, we performed sampling 20 times in the test set to measure the MSE and the variance (var) of the samples. Tab.~\ref{tab:T_exp} shows that reducing \( T \) improves accuracy at the cost of diversity in most cases. In the extreme case of \( T = 1 \), the method becomes deterministic. 

To explore models focused on diversity and accuracy, we develop two types of diffusion models: one with \( T=1000 \) and the other with \( T=1 \). Following the ablation study in Tab.~\ref{tab:tau_experiments}, we select \(\tau = (0.9, 1.2, 1.5)\) when \(T = 1000\) to prioritize diversity. Due to its gradual generation process, we name this model the \textbf{P}rogressive \textbf{D}iffusion \textbf{M}odel (PDM). However, this gradual generation process cannot be applied to the diffusion model with \( T=1 \). Moreover, \( T=1 \) limits the ability of the diffusion model to exploit its full potential over multiple timesteps~\cite{choi2022p2}. To this end, we define our new noise scheduler, as shown in Tab.~\ref{tab:sdm_noise}.

\begin{table}[h!]
\centering
\resizebox{0.96\columnwidth}{!}{
\begin{tabular}{|c||c|c|c|c|}
\hline
\(t\)             & \(t \mod 3 = 0\) & \(t \mod 3 = 1\) & \(t \mod 3 = 2\) & \(T-1\)\\ \hline
\(\bar{\alpha}^G_t\)         & 1                & 1                & 0     & 0  \\ \hline
\(\bar{\alpha}^M_t\)         & 1                & 0                & 1     & 0  \\ \hline
\(\bar{\alpha}^L_t\)         & 0                & 1                & 1     & 0  \\ \hline
\end{tabular}
}
\caption{\textbf{SDM noise scheduler. }Here, \(\bar{\alpha}^G_t\), \(\bar{\alpha}^M_t\), and \(\bar{\alpha}^L_t\) are the schedulers for geometry, material, and lighting, respectively.}
\label{tab:sdm_noise}
\end{table}

A value of 0 implies that pure Gaussian noise is added, while a value of 1 means that no noise is added. With our new noise scheduler, \( T \) is redefined as \( 3n+1 \) where \( n > 0 \). After the initial prediction at \( t = T-1 \), the noise is switched per channel, providing conditions for the next diffusion step. Taking into account this feature of the switchable noise scheduling, we name this model \textbf{S}witchable \textbf{D}iffusion \textbf{M}odel (SDM). The SDM is trained to predict the modality with an SNR of 0 at each step, allowing for more refined predictions by utilizing the previous step's predictions. 


\subsection{RGB-Guided Super Resolution}
We use a UNet as the RGB-guided super-resolution model, consisting of a shared encoder and five separate decoders. The encoder processes the RGB image into a dense feature map, which, along with the rendered diffuse \(\mathrm{I}_d\) and specular \(\mathrm{I}_s\) images from \(\mathbf{z}_0\), is fed into each decoder. Our RGB-guided Super-Resolution Model (SRM) is as follows:

\vspace{-2mm}
\begin{equation}\label{eqn:eq_sr}
\mathrm{N}, \mathrm{D}, \mathrm{A}, \mathrm{R}, \boldsymbol{f} = \text{SRM}(\mathbf{z}_0, \mathrm{I}, \mathrm{I}_d, \mathrm{I}_s),
\end{equation}

%% file: 4_experiments.tex
\section{Evaluation}  
In this section, we evaluate our method on both synthetic and real-world datasets. We first evaluate on a synthetic dataset, followed by inverse rendering and object insertion on a real-world spatially varying lighting dataset. Furthermore, we evaluated our method on the sub-tasks of inverse rendering, namely intrinsic decomposition and geometry prediction. More details on training, model architectures, and additional experimental results are provided in the Supplementary Materials.

\noindent{\bf Evaluation settings.}  
We used \(C = 128\), \(H = 240\), \(W = 320\), \(H_L = 60\), and \(W_L = 80\). For SDM, we set \(T = 4\). In all experiments, PDM generated 10 samples and the mean and variance of these samples were evaluated. 

\noindent{\bf Baselines.}  
We compared our method with state-of-the-art inverse rendering frameworks capable of per-pixel lighting inference, specifically those by \li\ and \zhu. Both methods were trained on the OpenRooms FF~\cite{choi2023mair} dataset. However, since the code for Zhu's lighting model has not been released, we used their pretrained lighting model, which was trained on the InteriorVerse dataset.

\begin{figure}[htb]
  \centering
  \includegraphics[width=\linewidth]{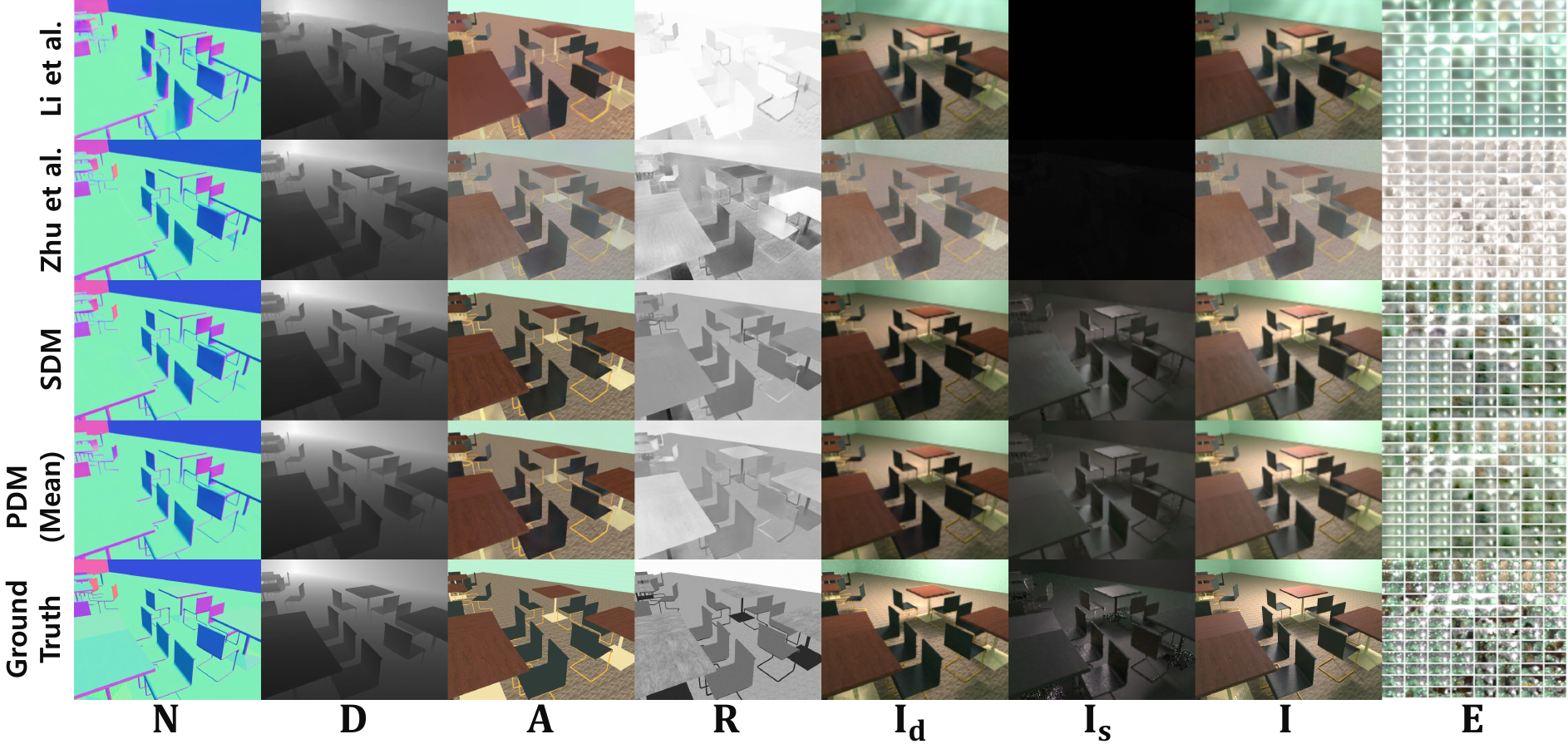}
  \caption{\textbf{Qualitative synthetic evaluation. }Only our method successfully decomposes specular radiance into material and lighting.}
   \label{fig:ir_syn}
\end{figure}

\begin{figure*}[t!]
  \centering
  \includegraphics[width=\linewidth]{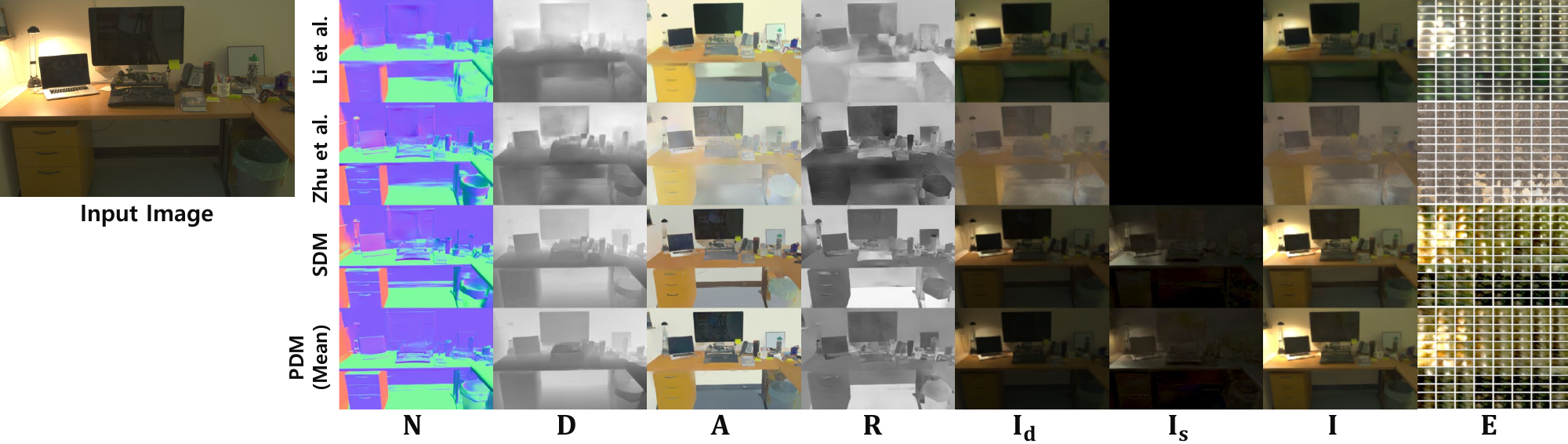}
  \caption{\textbf{Qualitative real-world evaluation. }Our method successfully recognizes shadows under the desk, achieving plausible geometry prediction and realistic re-rendering results. }
   \label{fig:ir_real}
\end{figure*}

\begin{figure*}[t!]
  \centering
  \includegraphics[width=\linewidth]{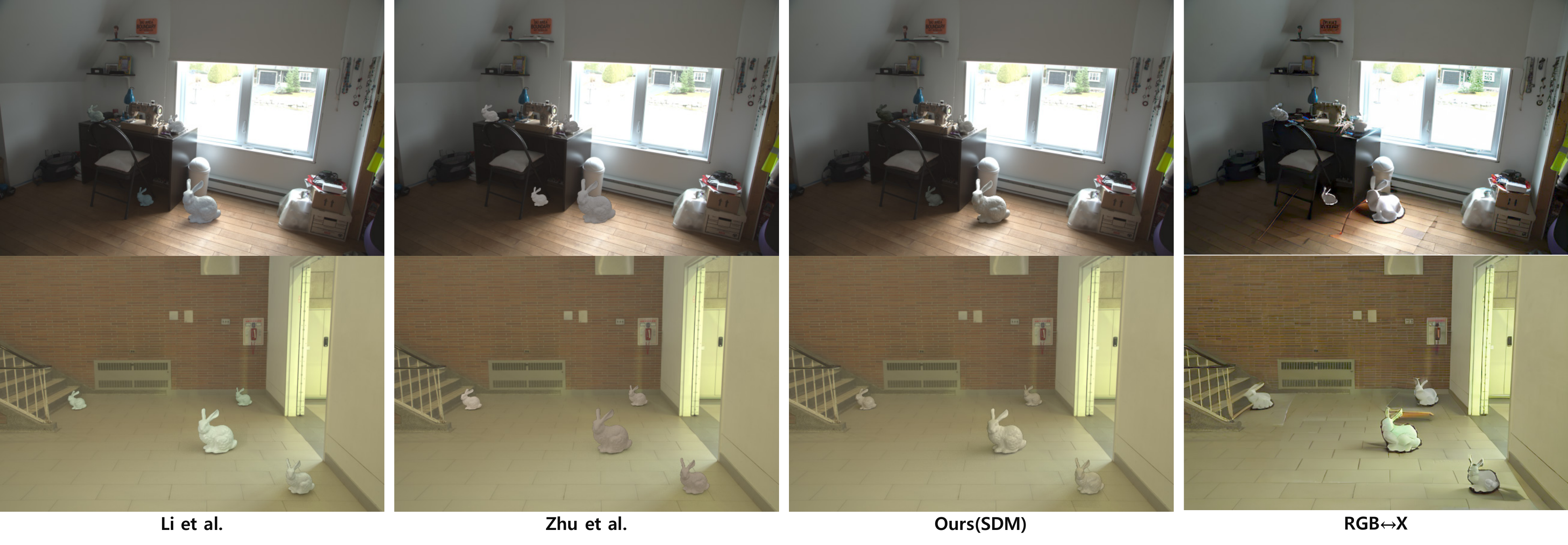}
  \caption{\textbf{Qualitative object insertion evaluation. }We accurately predicted the direction of light and shadows even in real-world images. Additionally, our rendered objects were closer to the ground truth in terms of hue and brightness compared to the baselines.
}
   \label{fig:oi_real}
\end{figure*}

\subsection{Evaluation on Synthetic Scenes}
Tab.~\ref{tab:compare} includes quantitative prediction results on the synthetic~\cite{choi2023mair} test dataset. Although our target is single-view, we include results from the multi-view baselines MAIR~\cite{choi2023mair} and MAIR++~\cite{mair++}. Our SDM, which prioritizes prediction accuracy, demonstrated superior performance compared to existing single-image baselines in all areas except roughness, and our PDM also demonstrated respectable performance comparable to the baselines. Notably, our lighting accuracy was comparable to that of MAIR, and the albedo accuracy outperformed even the multi-view methods. Additionally, it can be observed that both MAIR++ and our method, which are capable of ILR-based neural rendering, exhibit significantly superior re-rendering ($\mathrm{I}$) performance. Figure~\ref{fig:ir_syn} presents an illustrative example of this fact. Even in images with strong specular radiance, our SDM and PDM successfully decomposed and re-rendered the image in contrast to the failure of the baseline methods.

\begin{table}[ht]
\footnotesize
\centering
\resizebox{\columnwidth}{!}{
\begin{tabular}{|l||c|c|c|c|c|c|}
\hline
MSE($\times 10^{-2}$) & $\mathrm{N}$ $\downarrow$ & $\mathrm{D}$ $\downarrow$ & $\mathrm{A}$ $\downarrow$ & $\mathrm{R}$ $\downarrow$ & $\mathrm{E}$ $\downarrow$ & $\mathrm{I}$ $\downarrow$ \\ \hline
\li       &3.782     &0.135       &0.869       &6.496     &15.49   &0.491 \\ \hline
\zhu      &\underline{1.738}     &0.062        &0.528       &\textbf{4.208}     &*31.15  &*3.565  \\ \hline
SDM       &\textbf{1.502}     &\textbf{0.039}  &\textbf{0.349}       &4.969     &\textbf{13.35}   &0.156\\ \hline
PDM(Mean) &1.831     &0.047      &0.394        &4.819     &14.08   &\underline{0.150}    \\ \hline
PDM(Best) &1.782     &\underline{0.044}      &\underline{0.381}       &\underline{4.644}     &\underline{13.75}   &\textbf{0.146}   \\ \hline \hline
                     
MAIR~\cite{choi2023mair}      &1.010      &0.140       &0.639       &3.457     &13.79   &0.784       \\ \hline
MAIR++~\cite{mair++}    &1.010      &0.007       &0.478       &2.497     &11.61   &0.058       \\ \hline
\end{tabular}}
\caption{\textbf{Quantitative synthetic evaluation.} Bold text indicates the best performance among single-view methods, while underlined text represents the second-best. * denotes models trained on the InteriorVerse dataset.} 
\label{tab:compare}
\end{table}

\begin{figure*}[ht]
  \centering
  \includegraphics[width=\linewidth]{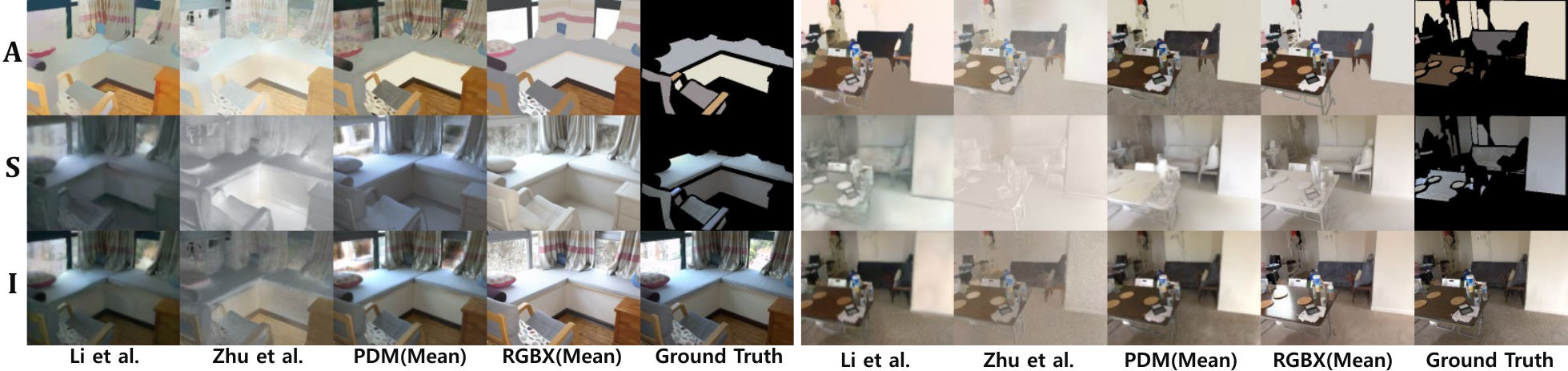}
  \caption{\textbf{Intrinsic decomposition.} Our method successfully decomposes the input image into albedo and shading. RGB$\leftrightarrow$X~\cite{zeng2024rgb}, leveraging its strong prior, achieves successful albedo prediction but struggles with realistic rendering, resulting in excessive brightness in the left sample and incorrect specular highlights in the right sample.}
  \label{fig:maw_vis}
\end{figure*}

\begin{figure}[ht]
  \centering
  \includegraphics[width=\linewidth]{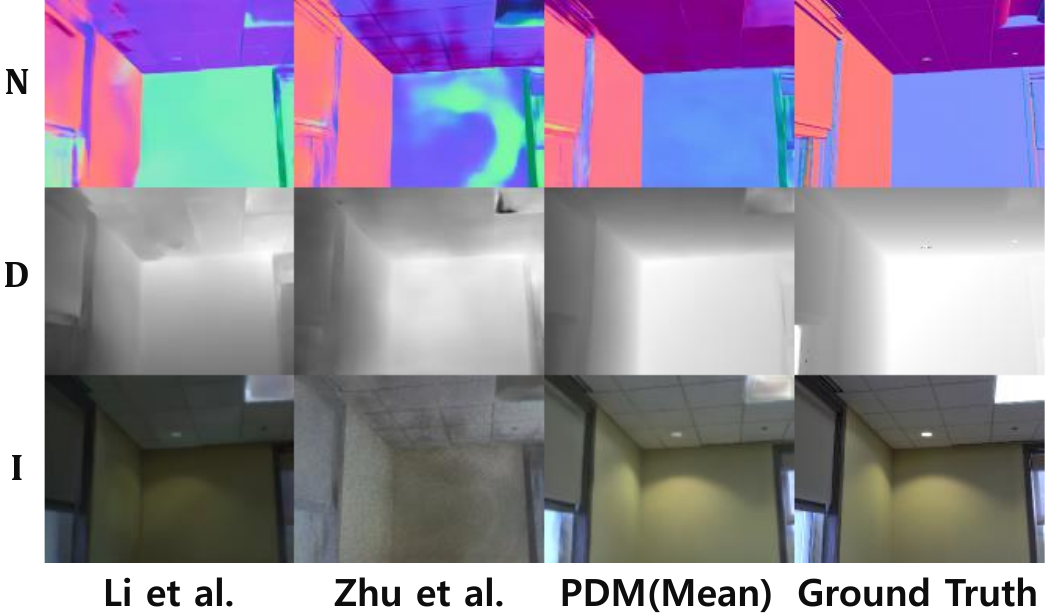}
  \caption{\textbf{Geometry prediction.} Our method achieves the most accurate prediction of the ceiling structure.}
  \label{fig:diode_vis}
\end{figure}

\subsection{Evaluation on Real-world Scenes}
\noindent{\bf Qualitative evaluation.} We evaluated the generalization capability of our method on real-world images from \garon. As shown in Fig.~\ref{fig:ir_real}, both SDM and PDM successfully separated the albedo even in the presence of strong shadows under the desk. \li\ partially removed the shadows but failed to predict the geometry, while \zhu\ accurately predicted the normal but failed to estimate the albedo.

\begin{table}[htb]
\centering
\resizebox{0.9\columnwidth}{!}{
\begin{tabular}{|c|c|c|c|}
\hline
\li &  \zhu     &SDM    &Ground Truth\\ \hline
0.178 &0.153    &0.322  &0.348   \\ \hline
\end{tabular}
}
\caption{\textbf{AMT user study on object insertion.} Our method was selected by nearly as many users as the ground truth.}
\label{tab:user_study}
\end{table}

\noindent{\bf Lighting Evaluation. }  
We also evaluated lighting performance through object insertion, using the object insertion tool from \li\ . For fairness, we excluded PDM and focused on evaluating only SDM. As shown in Fig.~\ref{fig:oi_real}, our method accurately predicts strong directional lighting, as well as the surrounding light's tone and shadows, resulting in the most realistic object insertions. In contrast, \zhu\ failed to predict the correct light direction, while \li\ struggles with realistic rendering due to a lack of high-frequency details in the environment map. The diffusion-based method RGB$\leftrightarrow$X~\cite{zeng2024rgb} relies on inpainting for object insertion, which compromises the fidelity of the inserted objects and leads to visible artifacts. Additionally, since it cannot handle environment maps, it struggled to accurately predict the direction of light. For a more quantitative evaluation, we conducted an AMT user study on object insertion, including the ground truth. The results in Tab.~\ref{tab:user_study} confirmed the superiority of our method. Users were asked: "Of the four images, please select the one that shows white rabbits blending naturally into the surrounding light." Our method was selected by users nearly as frequently as the ground truth.

\begin{table}[t]
\centering
\resizebox{\linewidth}{!}{
\begin{tabular}{|l||c|c|c|c|c|c|}
\hline
Method   & $\mathrm{A}$ $\downarrow$ & Hue $\downarrow$ & Texture $\downarrow$ & WHDR $\downarrow$ & $\mathrm{S}$ $\downarrow$ & $\mathrm{I}$ $\downarrow$ \\ \hline
\li       & 0.924  & 5.781  & 0.473 & 0.301 & 1.162 & 0.021  \\ \hline
\zhu      & 1.685  & 4.638  & 0.472 & 0.301 & 1.441 & 0.028  \\ \hline
SDM       & 0.705  & 5.056  & 0.459 & 0.293 & 1.181 & 0.006  \\ \hline
PDM(Mean) & 0.757  & 5.260  & 0.459 & 0.306 & 1.156 & 0.005 \\ \hline
PDM(Best) & 0.623  & 4.527  & 0.456 & 0.278 & \textbf{1.113} & \textbf{0.005} \\ \hline
RGB$\leftrightarrow$X~\cite{zeng2024rgb}(Mean) & 0.602  & 3.690  & 0.448 & 0.168 & 1.261 & 0.055 \\ \hline
RGB$\leftrightarrow$X~\cite{zeng2024rgb}(Best) & \textbf{0.331}  & \textbf{3.091}  & \textbf{0.441} & \textbf{0.146} & 1.196 & 0.052  \\ \hline
\end{tabular}
}
\vspace{-2.5mm}
\caption{\textbf{Intrinsic decomposition evaluation on MAW~\cite{wu2023maw}.} Bold text indicates the best performance.}
\label{tab:compare_maw}
\end{table}

\subsection{Evaluation on Sub-tasks}
\noindent{\bf Intrinsic decomposition. }We evaluated our method on intrinsic decomposition using the recently introduced MAW~\cite{wu2023maw} dataset, which provides more refined annotations for 888 images, allowing more quantitative comparisons than the WHDR metric used in the IIW~\cite{iiw} dataset. Tab.~\ref{tab:compare_maw} presents comparisons for MSE for $\mathrm{A}$, $\mathrm{S}$, and $\mathrm{I}$, along with metrics for hue, texture, and WHDR for $\mathrm{A}$. The hue metric evaluates the chromaticity component, while the texture metric is based on LPIPS perceptual similarity. Our method generally outperformed existing deterministic methods~\cite{cis2020, zhu2022montecarlo}. RGB$\leftrightarrow$X~\cite{zeng2024rgb}, which learns strong priors from diverse large datasets, excels in albedo inference; however, due to its reliance on diffusion-based rendering, it reconstructed the input images with the lowest accuracy. In contrast, our method, trained only on OpenRooms FF, achieved the highest performance in shading estimation and most faithfully reproduced input images, demonstrating high fidelity. This can also be observed in Fig.~\ref{fig:maw_vis}. In the left sample, RGB$\leftrightarrow$X~\cite{zeng2024rgb} fails to predict the shading intensity as it does not account for HDR lighting. Similarly, in the right sample, RGB$\leftrightarrow$X~\cite{zeng2024rgb} introduces specular highlights on the desk that are absent in the input image, resulting in a loss of fidelity. In contrast, our method predicts spatially consistent albedo and accurately estimates brightness, enabling a more realistic reproduction of the input image.


\noindent{\bf Geometry prediction. }Tab.~\ref{tab:compare_diode} includes the geometry prediction evaluation on the DIODE~\cite{vasiljevic2019diode} indoor dataset. Similarly to the intrinsic decomposition task, both our PDM and SDM demonstrate superior or comparable performance to the baselines. Fig.~\ref{fig:diode_vis} provides qualitative examples, demonstrating that our method exhibits strong generalization capability in geometry prediction.


\begin{table}[ht]
\centering
\resizebox{\linewidth}{!}{
\begin{tabular}{|l|c|c|c|c|c|}
\hline
Metric & \li & \zhu & SDM & PDM(Mean) & PDM(Best) \\ \hline
$\mathrm{N}$ $\downarrow$ & 0.196 & 0.171 & 0.147 & 0.152 & \textbf{0.144} \\ \hline
$\mathrm{D}$ $\downarrow$ & 0.059 & 0.057 & 0.057 & 0.055 & \textbf{0.050} \\ \hline
$\mathrm{I}$ $\downarrow$ & 0.015 & 0.013 & 0.003 & \textbf{0.002} & \textbf{0.002} \\ \hline
\end{tabular}
}
\vspace{-2.5mm}
\caption{\textbf{Geometry prediction evaluation on DIODE~\cite{vasiljevic2019diode}.} Bold text indicates the best performance.}
\label{tab:compare_diode}
\end{table}

\begin{figure}[htb]
  \centering
  \includegraphics[width=\linewidth]{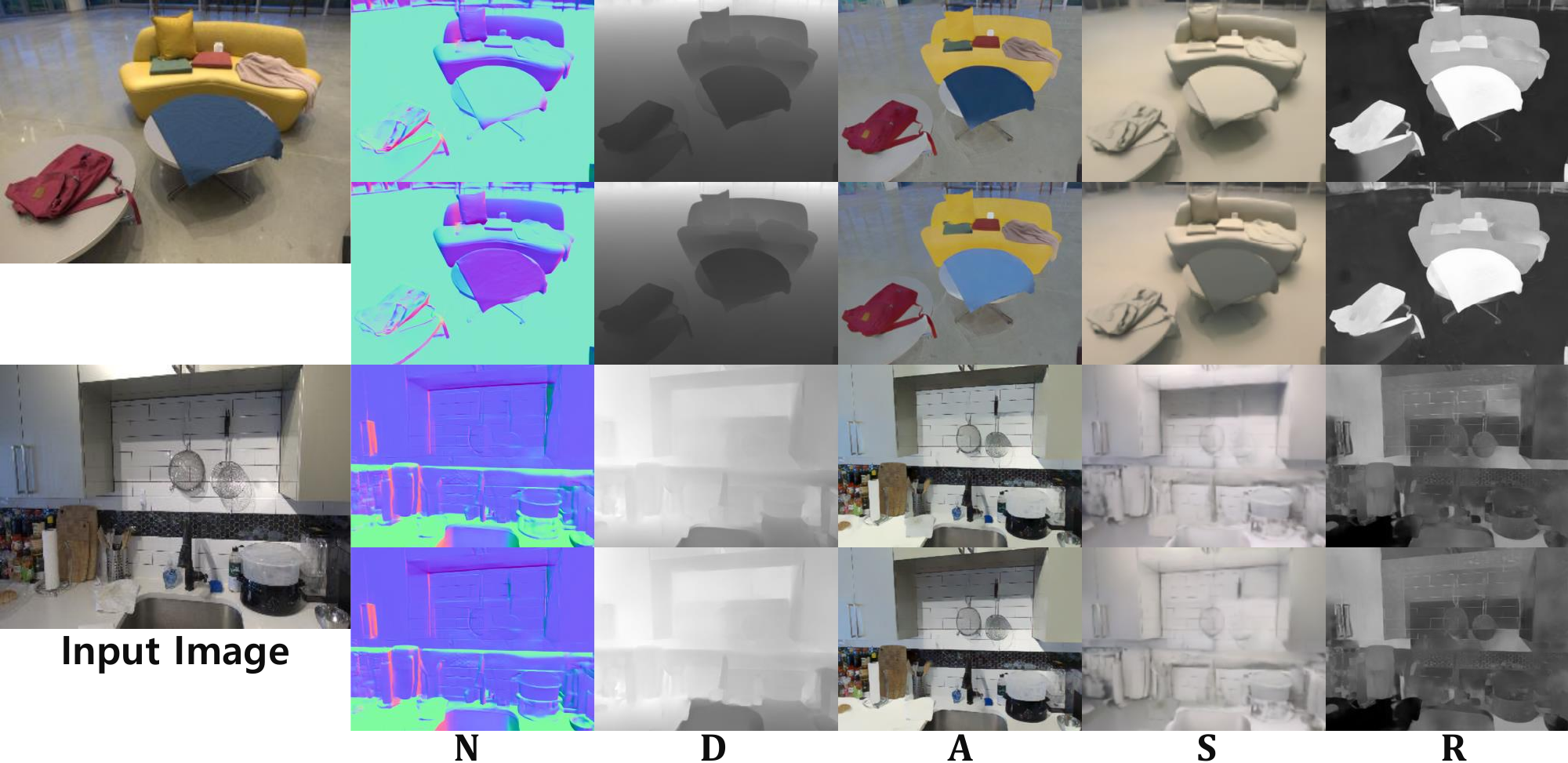}
  \caption{\textbf{Modality dependencies. }Two samples of $\mathrm{N}$, $\mathrm{D}$, $\mathrm{A}$, $\mathrm{S}$, and $\mathrm{R}$ predicted by PDM.}
   \label{fig:pdm_diversity}
\end{figure}

\subsection{Modality Dependencies}  
To generate a reasonable solution for inverse rendering, it is crucial to account for the dependencies between modalities. For instance, if a material is predicted to be darker, the lighting should be predicted brighter to compensate for the input image. However, existing diffusion-based methods~\cite{luo2024intrinsicdiffusion, zeng2024rgb, kocsis2023intrinsic} infer each modality independently, making it difficult to capture such dependencies. In contrast, our PDM generates all modalities simultaneously, enabling it to account for these relationships. Fig.~\ref{fig:pdm_diversity} illustrates this with two different samples generated by PDM. For the center table in the top image, the first sample predicts the normal pointing upward with bright shading, resulting in a darker albedo. Conversely, the second sample predicts the normal facing the camera, producing darker shading and a brighter albedo. Additionally, for the area under the cabinet in the bottom image, the first sample interprets it as a shadow and predicts a bright albedo, while the second sample treats it as part of the albedo and generates bright shading.

\subsection{Ablations}
\begin{figure}[htb]
  \centering
  \includegraphics[width=\linewidth]{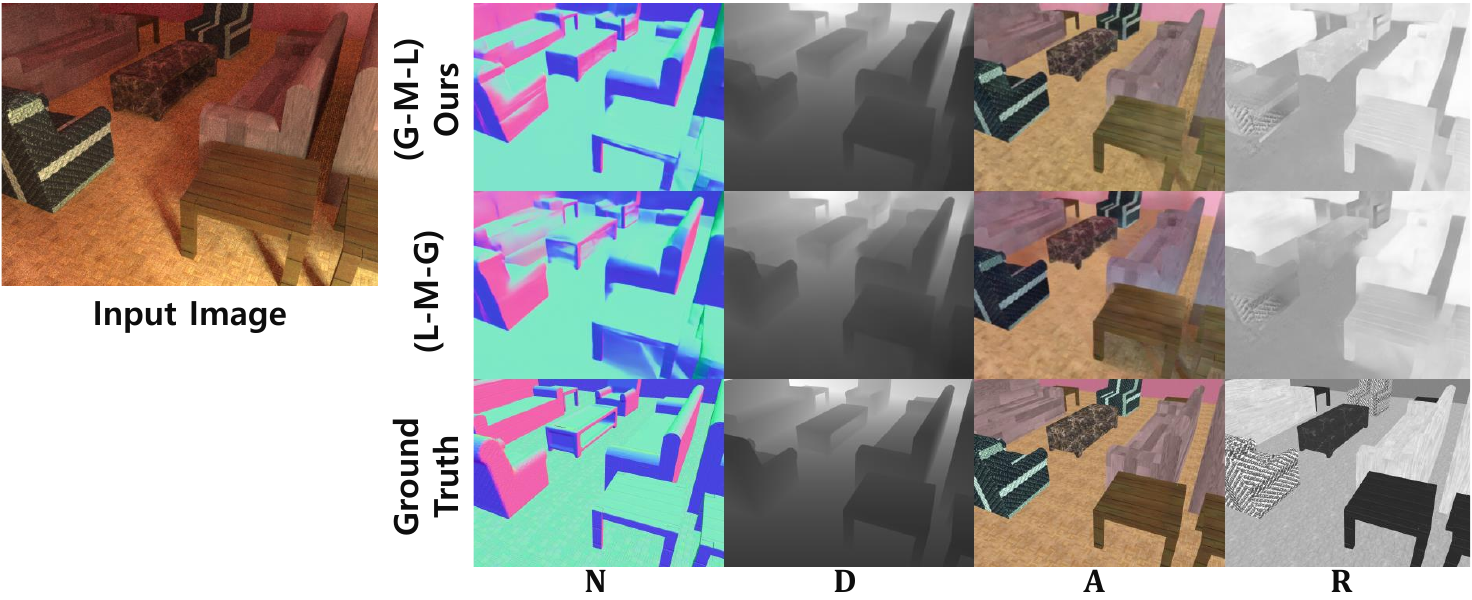}
 \caption{\textbf{Comparison of modality generation order.} Generating in the order of geometry, material, and lighting helps effectively infer shadows.}

   \label{fig:noise_abl}
\end{figure}

\noindent{\bf Modality generation order. } To explore the optimal modality generation order, we conducted an ablation study on the modality generation sequence. There are no restrictions on \( \tau \); however, we selected three values: \( \tau = 0.9, 1.2, \) and \( 1.5 \). These values were applied separately to \( \tau_G \) (geometry), \( \tau_M \) (material), and \( \tau_L \) (lighting), and the models were trained independently. The performance was then evaluated on the test dataset. Tab.~\ref{tab:tau_experiments} shows that this channel-wise noise scheduling can significantly impact the inverse rendering quality. In particular, \(\tau_G, \tau_M, \tau_L = (0.9, 1.2, 1.5)\) produces the best quality, aligning with our intuition. In contrast, generating in reverse order results in a detrimental effect. Fig.~\ref{fig:noise_abl} provides examples of these two cases. The reverse order (L-M-G) exhibits artifacts in areas with strong shadows, whereas our method produces plausible results.

\begin{table}[htb]
\centering
\resizebox{\columnwidth}{!}{
\begin{tabular}{|c|c|c||c|c|c|c|c|c|}
\hline
$\tau_G$ &$\tau_M$ &$\tau_L$ &$\mathrm{N}_L \downarrow$ &$\mathrm{D}_L \downarrow$ &$\mathrm{A}_L \downarrow$ &$\mathrm{R}_L \downarrow$ &$\mathrm{E}_L \downarrow$ &$\mathrm{I}_L \downarrow$  \\ \hline
1.0      &1.0      &1.0      &2.140     &0.062    &0.403    &5.754     &12.19   &0.163    \\ \hline
0.9      &1.2      &1.5      &\textbf{2.024}     &\textbf{0.059}    &\textbf{0.376}    &\textbf{5.176}     &\textbf{11.94}   &\textbf{0.159}    \\ \hline
0.9      &1.5      &1.2      &2.104     &0.060    &0.390    &5.723     &12.15   &0.165    \\ \hline
1.2      &0.9      &1.5      &2.073     &0.060    &0.388    &5.394     &12.20   &0.161    \\ \hline
1.2      &1.5      &0.9      &2.202     &0.064    &0.411    &5.882     &12.33   &\underline{0.167}    \\ \hline
1.5      &0.9      &1.2      &2.211     &0.064    &0.405    &5.664     &12.36   &0.160    \\ \hline
1.5      &1.2      &0.9      &\underline{2.245}     &\underline{0.066}    &\underline{0.414}    &\underline{5.986}     &\underline{12.53}   &0.166    \\ \hline
\end{tabular}
}
\caption{\textbf{Ablation of the generation order.} The best values are highlighted in \textbf{bold}, while the worst values are \underline{underlined}. }
\label{tab:tau_experiments}
\end{table}

\noindent{\bf Iterative Prediction of SDM. }  We conducted an ablation study to evaluate the benefits of iterative prediction for SDM. In Tab.~\ref{tab:abl_SDM}, the improvements at \( T=4 \) and \( T=7 \) over \( T=1 \) indicate that iterative prediction utilizes previously predicted modalities to inform subsequent predictions. However, the performance drop at \( T=7 \) compared to \( T=4 \) suggests that excessive iterations do not always yield better results.

\begin{table}[t]
\centering
\small 
\resizebox{0.9\linewidth}{!}{
\begin{tabular}{|c||c|c|c|c|c|c|}
\hline
$T$ & $\mathrm{N}$ $\downarrow$ & $\mathrm{D}$ $\downarrow$ & $\mathrm{A}$ $\downarrow$ & $\mathrm{R}$ $\downarrow$ & $\mathrm{E}$ $\downarrow$ & $\mathrm{I}$ $\downarrow$ \\ \hline
1 & 1.733 & 0.060 & 0.318 & 5.837 & 11.24 & \textbf{0.132} \\ \hline 
4 & \textbf{1.578} & 0.052 & \textbf{0.288} & \textbf{5.743} & \textbf{10.96} & 0.137 \\ \hline 
7 & 1.630 & \textbf{0.050} & 0.292 & 5.759 & 10.97 & 0.139 \\ \hline
\end{tabular}
}
\caption{\textbf{Ablation of SDM iterative prediction.} Bold text indicates the best performance.} 
\label{tab:abl_SDM}
\end{table}

\section{Conclusion}
We tackled inverse rendering with two opposing goals: attaining diversity and ensuring accuracy. To address these objectives, we proposed a diffusion-based framework, in which channel-wise noise scheduling allowed each diffusion model to better understand the correlations between different modalities. 


\noindent{\bf Limitations and future work. }
Our approach has the limitation of requiring two separate models; thus, integrating them into a unified model would enhance usability. 
We explored re-rendering error as an alternative metric to overcome the constraint of depending only on the predicted variance of PDM for estimating inverse rendering uncertainty, but it could be imprecise. Further research on the handling of uncertainty could improve the complementarity of SDM and PDM. Moreover, our method refrains from exploiting the priors of pretrained LDMs. Using those priors to process per-pixel environment maps might enhance the performance and expand the applications of LDMs.

\noindent\textbf{Acknowledgements.}
This work was supported by Institute of Information \& communications Technology Planning \& Evaluation (IITP) grant 
funded by the Korea government(MSIT)(No.RS-2023-00227592, Development of 3D Object Identification Technology Robust to Viewpoint Changes)

%% file: 5_supplementary.tex
\maketitlesupplementary
\appendix
\setcounter{page}{1}
In this supplementary material, we first demonstrate the complementary nature of PDM and SDM, highlighting the importance of considering both accuracy and diversity in inverse rendering (\cref{sec:supp:complementarity}). Second, we provide the implementation details (\cref{sec:supp:implementation}). Next, we present the application results (\cref{sec:supp:applications}), and finally, we provide additional experimental results (\cref{sec:supp:experiments}).

\section{Validation of the Complementarity}\label{sec:supp:complementarity}  
\begin{figure}[ht]
  \centering
  \includegraphics[width=\linewidth]{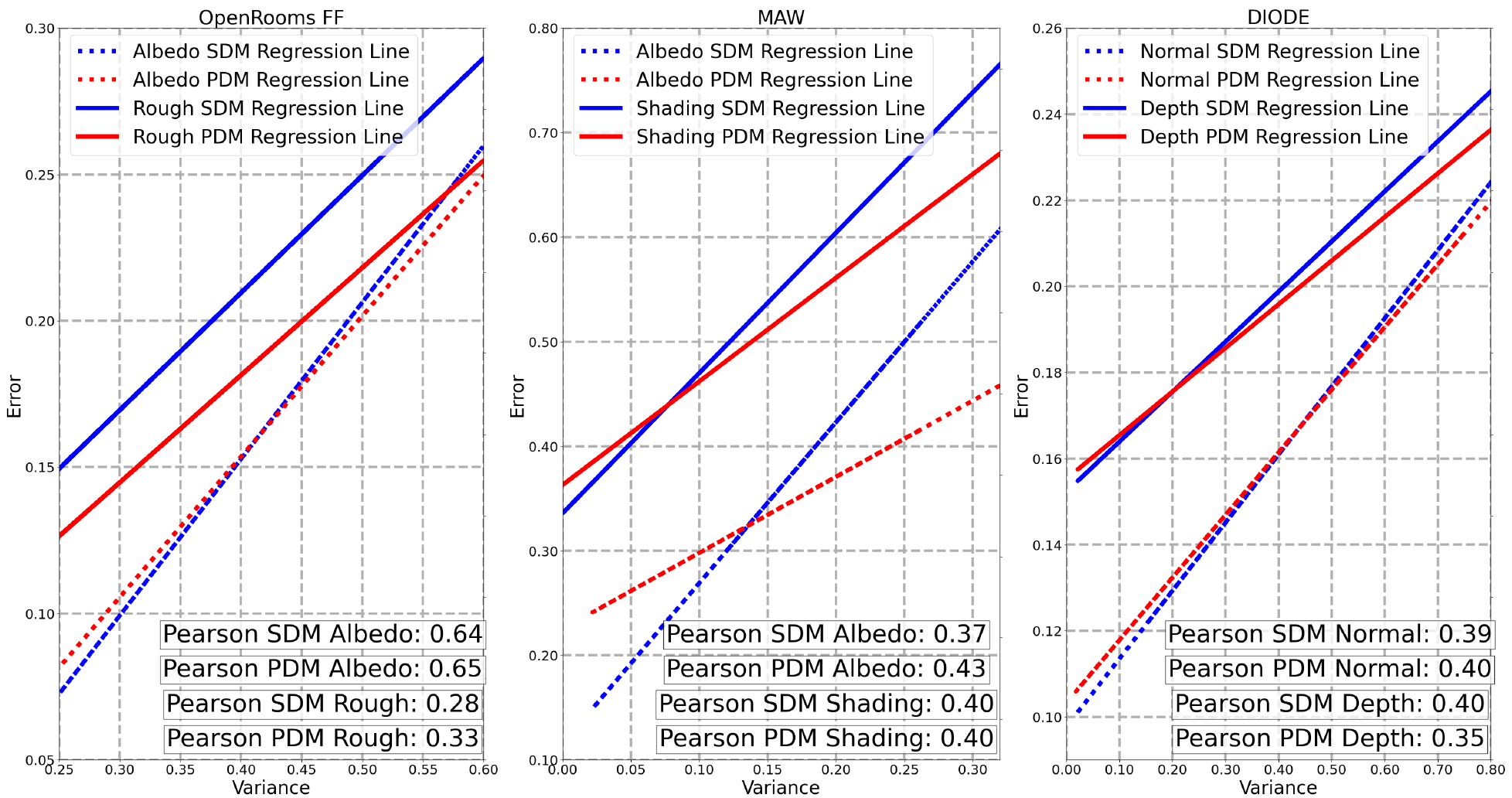}
   \caption{\textbf{Correlation of variance(x-axis) and error(y-axis).} The error of PDM was calculated from the mean predictions. The blue line represents SDM, while the red line represents PDM.}
  \label{fig:acc_var}
\end{figure}

As shown in Figs.~\ref{fig:supp_ir_indoor}, \ref{fig:supp_ir_real}, \ref{fig:supp_ir_maw}, and \ref{fig:supp_ir_diode}, our PDM exhibits high variance in ambiguous regions, making it a useful measure for assessing uncertainty in inverse rendering. We leveraged this characteristic to validate the complementarity between PDM and SDM. In Fig.~\ref{fig:acc_var}, we measured the variance and error of PDM across datasets and conducted linear regression to analyze the correlation between error and variance. Each graph includes the Pearson correlation coefficient in the lower right corner. As expected, a higher variance led to a higher error (lower accuracy) across all datasets. This is confirmed by the positive Pearson coefficients, indicating that PDM presents diverse solutions for ambiguous regions. Furthermore, SDM (blue line) excels in low-variance images, while PDM (red line) outperforms in ambiguous, high-variance images. This fact demonstrates the complementary nature of PDM and SDM, emphasizing the importance of incorporating both accuracy and diversity in inverse rendering. For the roughness map of OpenRooms FF, due to the inherent uncertainty of the modality, PDM consistently outperformed SDM.

For a more thorough validation, in Tab.~\ref{tab:variance}, we measured the average variance of $\mathrm{N}$, $\mathrm{D}$, $\mathrm{A}$, and $\mathrm{R}$ for each dataset. OpenRooms FF, where SDM outperforms (see Tab.~\ref{tab:compare}), exhibits low variance, indicating that it contains a large number of straightforward images. In contrast, the high variance of MAW and DIODE reflects ambiguity in complex real-world images. As shown in Tab.~\ref{tab:compare_maw} and Tab.~\ref{tab:compare_diode}, PDM proves to be a useful choice for handling such ambiguous images.


\begin{table}[htb]
\centering
\resizebox{0.9\columnwidth}{!}{
\begin{tabular}{|l||c|c|c|}
\hline
Dataset &OpenRooms FF~\cite{choi2023mair} & MAW~\cite{wu2023maw} & Diode~\cite{vasiljevic2019diode} \\ \hline
variance&0.039                            & 0.179                & 0.184                             \\ \hline
\end{tabular}
}
\caption{\textbf{PDM sample variance ($\times 10^{-2}$) for each dataset.}}
\label{tab:variance}
\end{table}

\section{Implementation Details}\label{sec:supp:implementation}  
All experiments were performed on eight A5000 GPUs and AdamW~\cite{adamw} optimizer was used for all models. Next, we provide the implementation details for each model. Each model was trained separately and subsequently frozen.

\subsection{Implicit Lighting Representation}
The input per-pixel environment map $\mathrm{E}$ is flattened on a pixel basis, followed by the application of \texttt{log1p}. The encoder consists of 7 layers, each with 64 hidden units, followed by a 1D BatchNorm. The feature vector $\boldsymbol{f}$ is constrained to the range \([-1, 1]\) by a \texttt{tanh} activation and has a dimensionality of 96. The decoders \(\text{Decoder}_E\), \(\text{Decoder}_S\), and \(\text{Decoder}_I\) are each composed of 3, 3, and 6 layers with 128 hidden units, respectively, without any normalization layers. Furthermore, to improve the robustness of \(\text{Decoder}_S\), random roughness values were generated and used during training. The loss function consists of a log-space \(L_2\) loss for \(\mathrm{E}\), \(\mathrm{S}\), \(\mathrm{I}_S\), \(\mathrm{I}_S\) (random), and a standard \(L_2\) loss for \(\mathrm{I}\). The model was trained with a batch size of 256 for 30 epochs, which took approximately one day to complete. These encoders and decoders are jointly trained and are subsequently frozen.

\subsection{Diffusion-based Inverse Rendering}
The DM was based on the \texttt{UNet2DConditionModel} from Diffusers~\cite{diffusers}, which consists of 586M parameters. We applied classifier-free guidance~\cite{ho2022classifier} with a 0.05 probability and rescaled the noise schedule following Lin \emph{et al.}~\cite{lin2024common}. The training was carried out with a batch size of 512 for 150 epochs, which took approximately 2 days to complete. After training, PDM used 10 DDIM~\cite{ddim} inference steps.

\subsection{RGB-Guided Super Resolution}
The SRM's encoder includes an encoder identical to the DenseNet~\cite{densenet} structure used for \(\mathrm{I}\), which produces dense feature maps. In addition, a separate encoder for \(\mathbf{z}_0\) is provided. This encoder consists of a single DenseBlock, a component of DenseNet, and its output is later combined with the dense feature maps as input to the modality-specific decoders. Each decoder also receives the corresponding low-resolution modality as input. The decoders follow the same structure as the existing baselines~\cite{cis2020}. The SRM was trained with a batch size of 128 for 50 epochs, taking approximately 2 days to complete. The loss function consists of an \(L_2\) loss for \(\mathrm{N}\), a scale-invariant log-space \(L_2\) loss for \(\mathrm{D}\), a scale-invariant \(L_2\) loss for \(\mathrm{A}\), an \(L_2\) loss for \(\mathrm{R}\), and an \(L_1\) loss for \(\boldsymbol{f}\). 

\begin{figure}[ht]
  \centering
  \includegraphics[width=\linewidth]{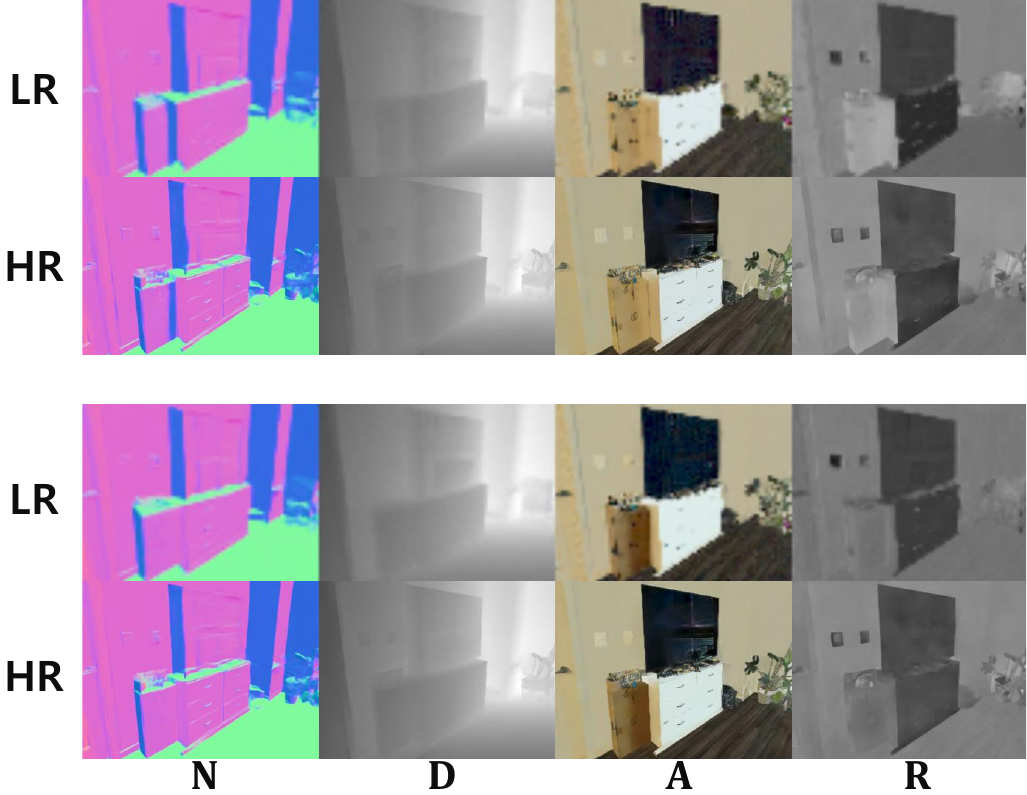}
   \caption{Performance Analysis of SRM. LR refers to the low-resolution sample, while HR denotes the high-resolution sample processed by SRM.}
  \label{fig:supp_sr}
\end{figure}

\noindent{\bf Performance Analysis of SRM. }  
For PDM, there was a tendency for slight residual noise to remain even after training. We analyzed whether SRM effectively removes this noise while preserving the identity of the samples generated by PDM. Fig.~\ref{fig:supp_sr} presents the results of this analysis. Both the top and bottom low-resolution samples in Fig.~\ref{fig:supp_sr} contain slight residual noise. However, in both cases, SRM successfully removes this noise and performs upsampling while maintaining the identity of each modality.

\section{Applications}\label{sec:supp:applications}  
\subsection{Material Editing}  
\begin{figure}[ht]
  \centering
  \includegraphics[width=\linewidth]{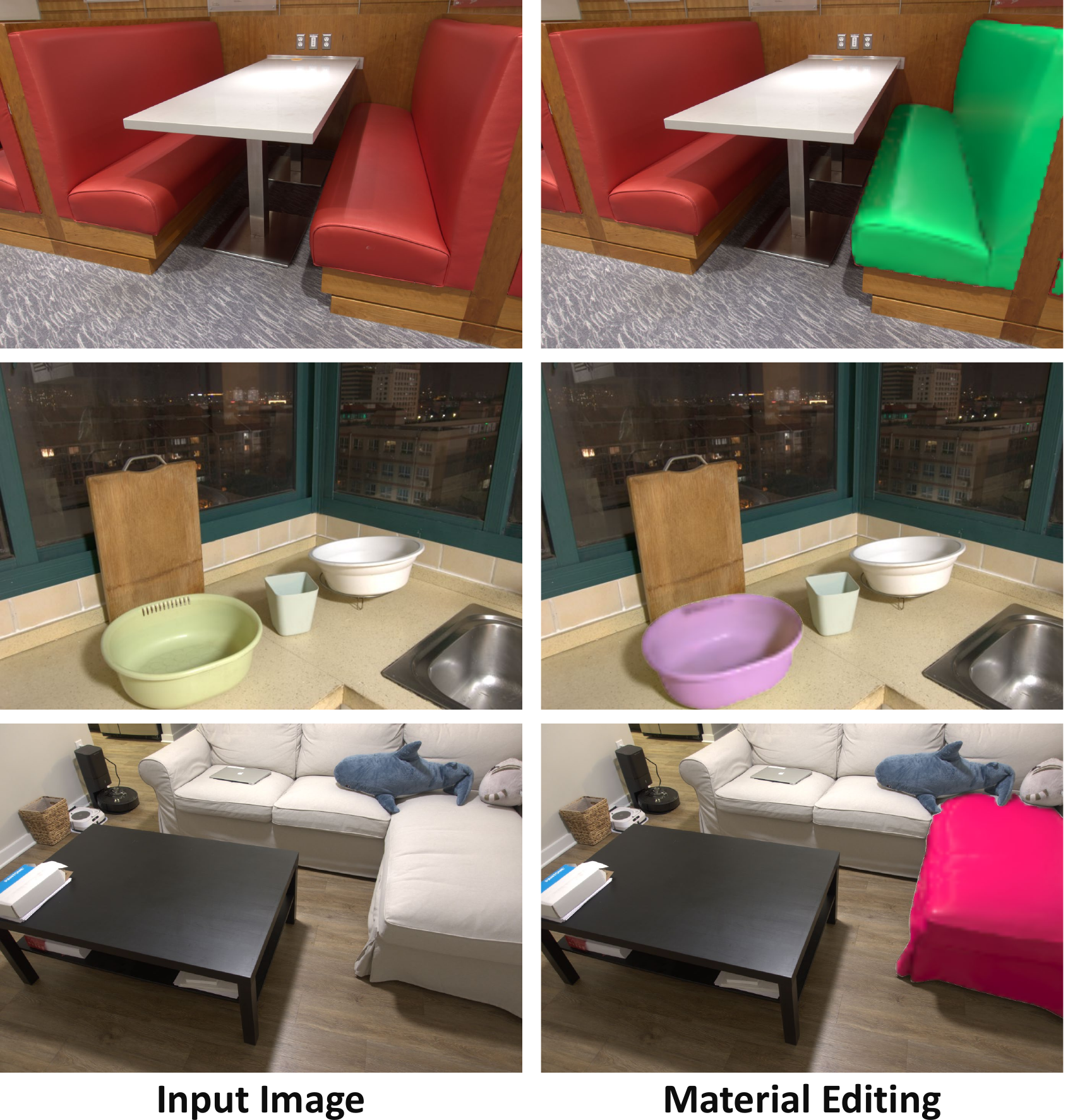}
   \caption{Material editing results.}
  \label{fig:supp_mat_edit}
\end{figure}

Inverse rendering, which decomposes an image into geometry, material, and lighting, enables applications such as material editing. In particular, ILR-based neural rendering allows for realistic rendering without the need for an external renderer. Fig.~\ref{fig:supp_mat_edit} demonstrates this capability. In each image, we modified the albedo to green, light purple, and pink, respectively, and for the third image, we reduced the roughness to introduce additional specularity. In all results, the spatially-varying lighting of the input image is faithfully reproduced, and the materials are realistically modified.

\subsection{Object Insertion}  
\begin{figure*}[t]
  \centering
  \includegraphics[width=1.0\linewidth]{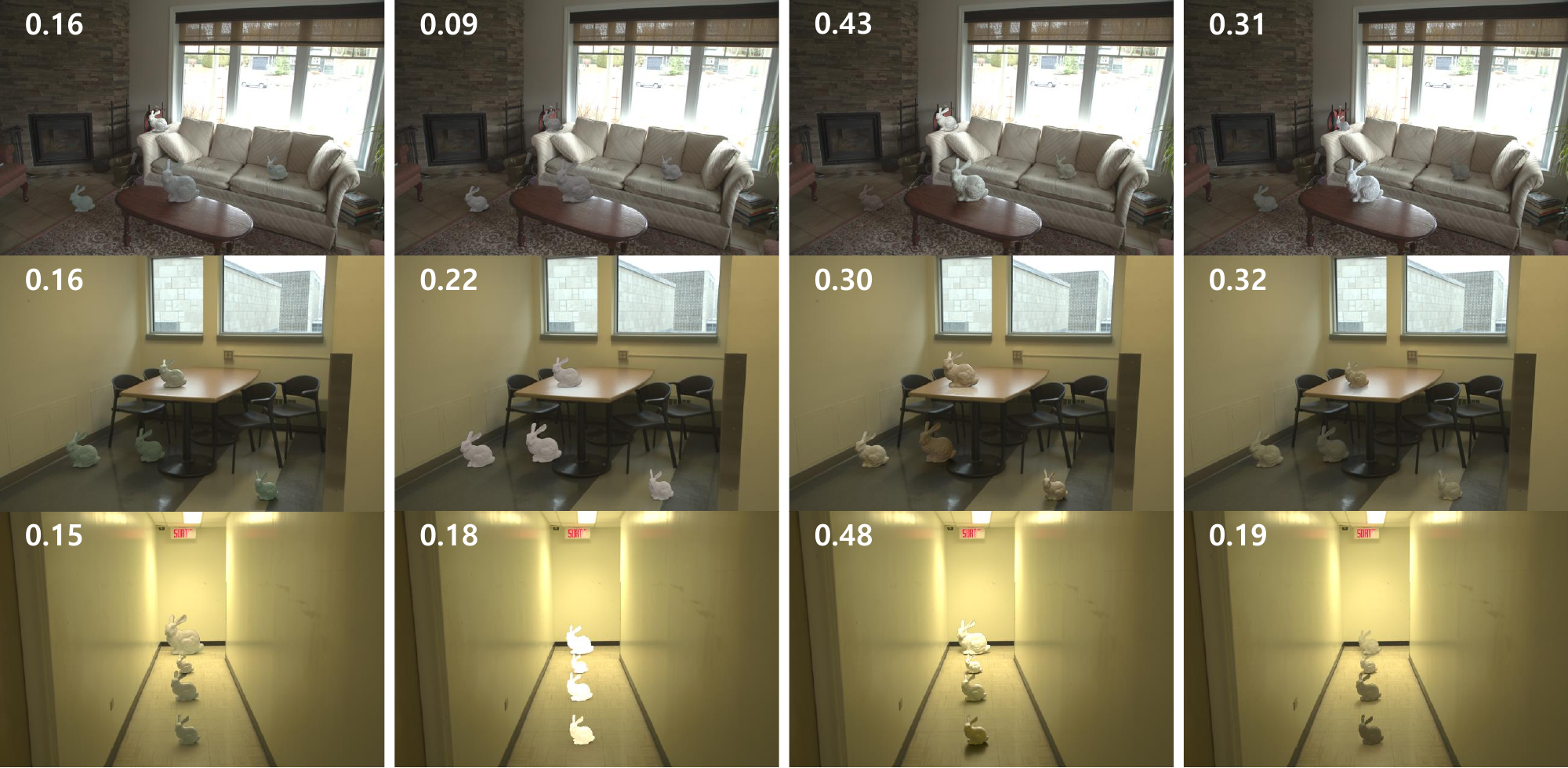}
  \includegraphics[width=1.0\linewidth]{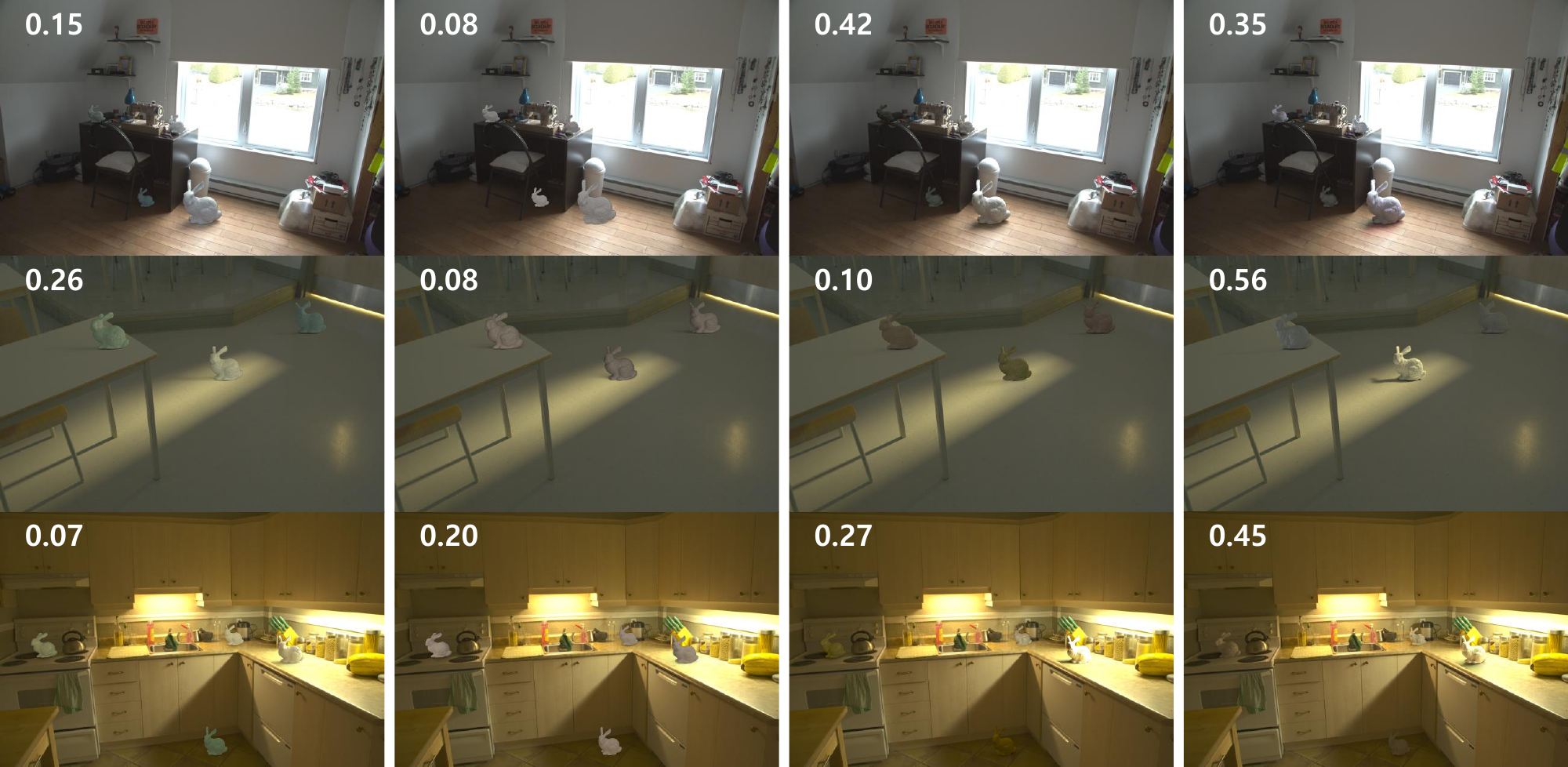}
  \includegraphics[width=1.0\linewidth]{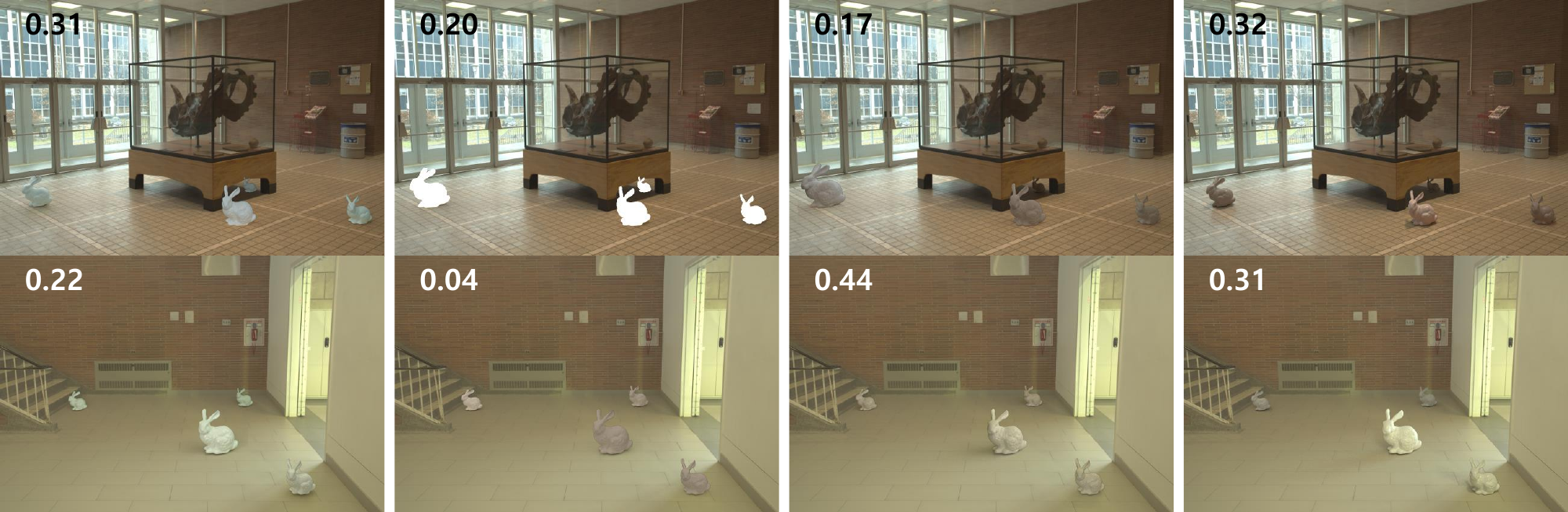}
   \caption{Additional object inversion results on spatially-varying lighting dataset.~\cite{garon19}.}
   \label{fig:supp_oi1}
\end{figure*}

\begin{figure*}[t]
  \centering
  \includegraphics[width=1.0\linewidth]{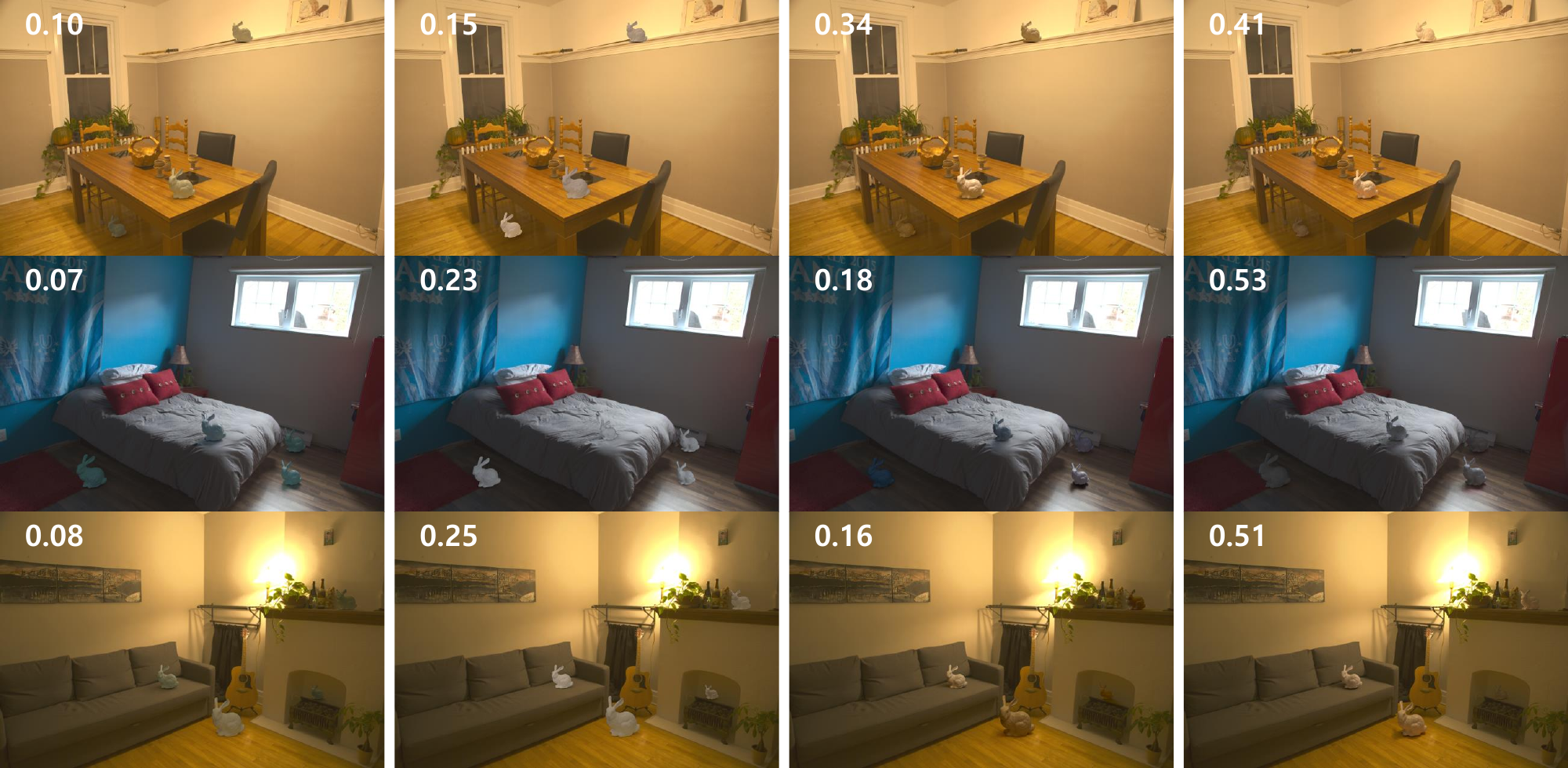}
  \includegraphics[width=1.0\linewidth]{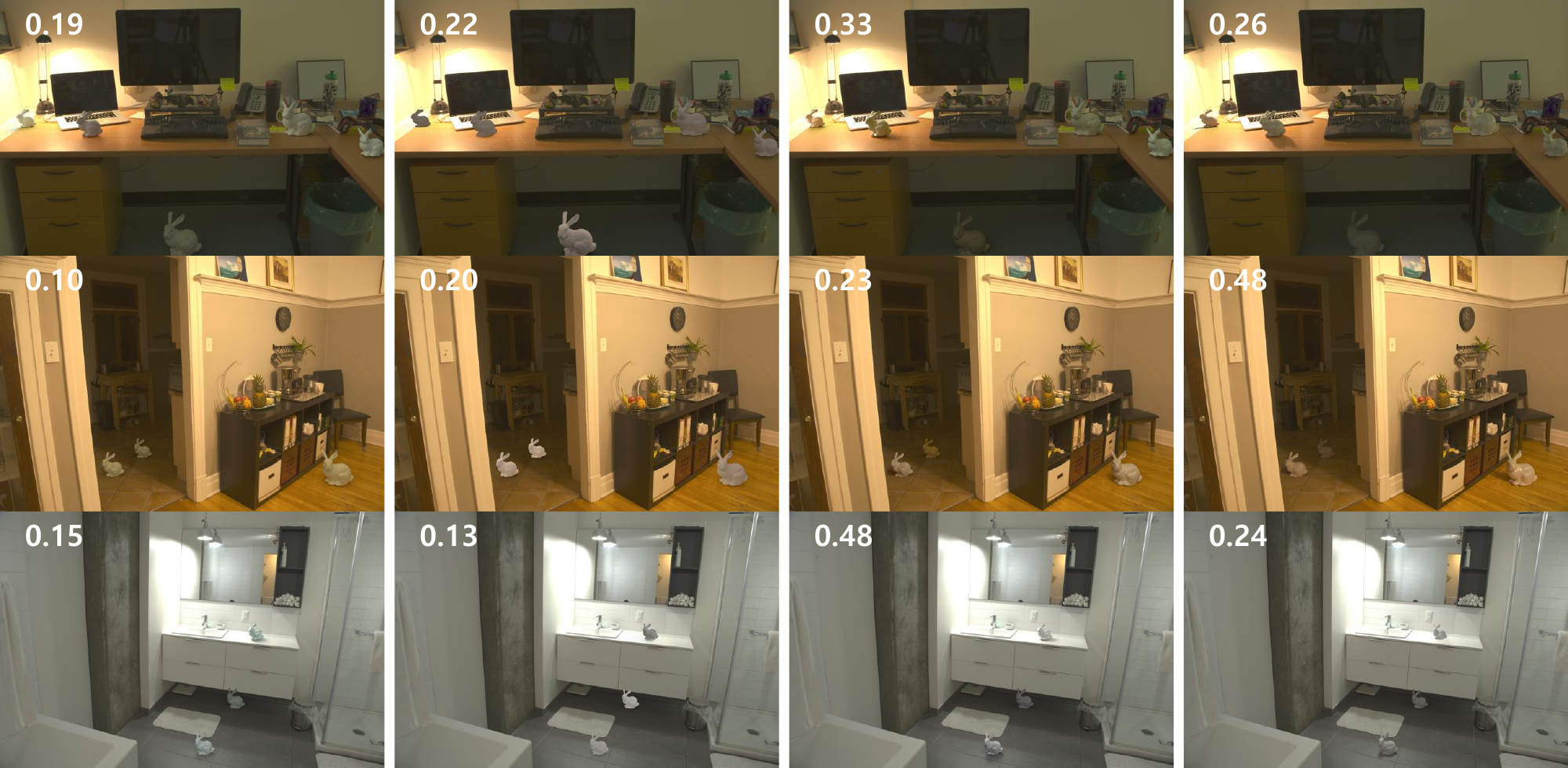}
  \includegraphics[width=1.0\linewidth]{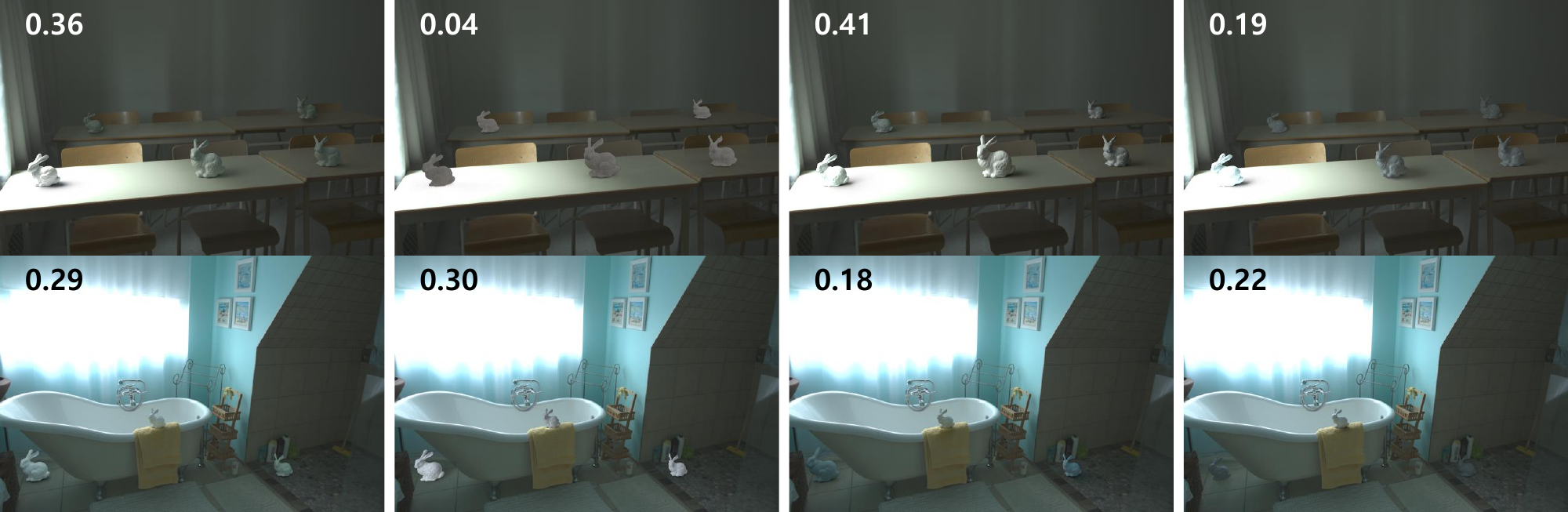}
   \caption{Additional object inversion results on spatially-varying lighting dataset.~\cite{garon19}.}
   \label{fig:supp_oi2}
\end{figure*}

\begin{figure*}[t]
  \centering
  \includegraphics[width=1.0\linewidth]{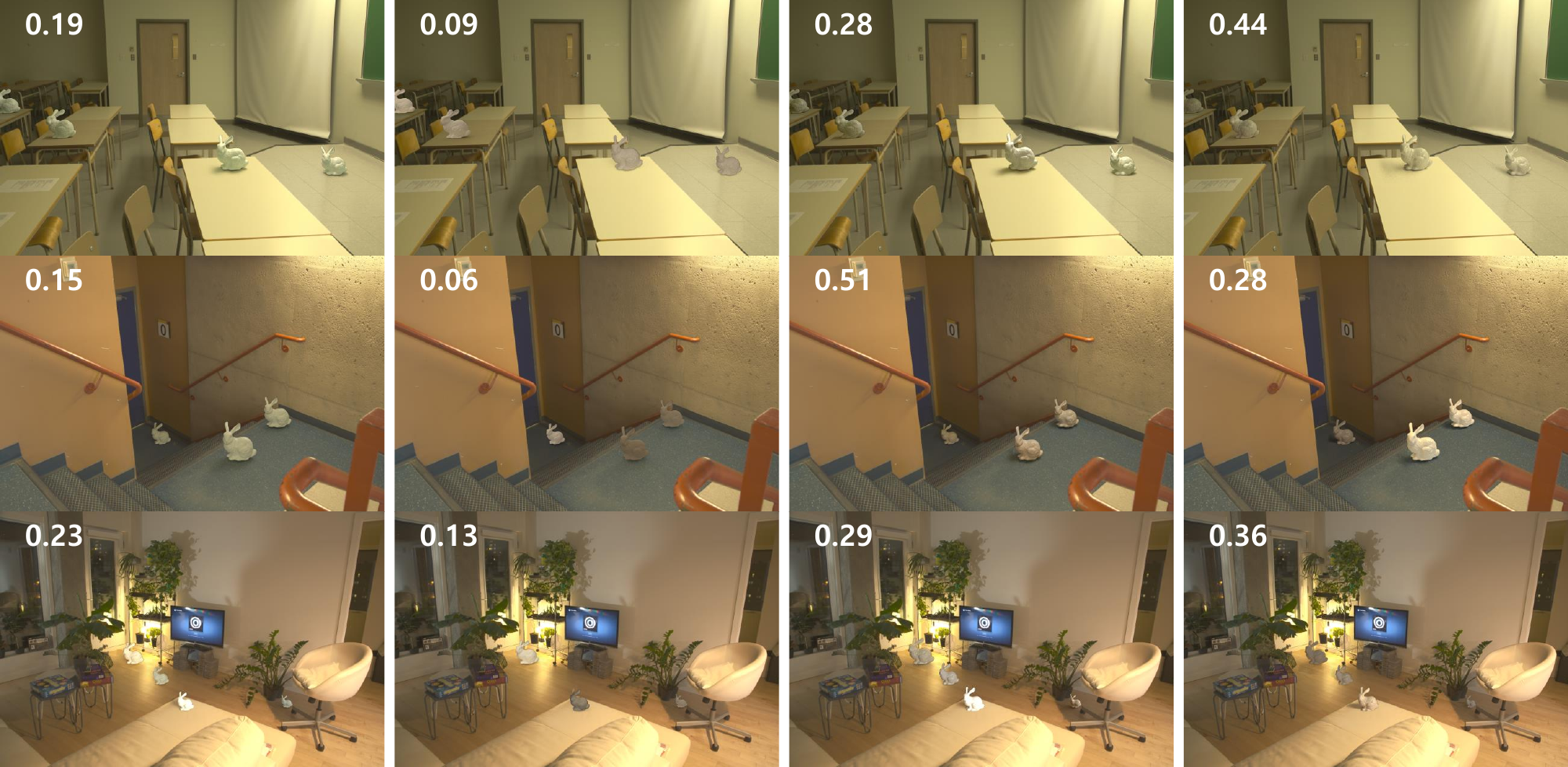}
  \includegraphics[width=1.0\linewidth]{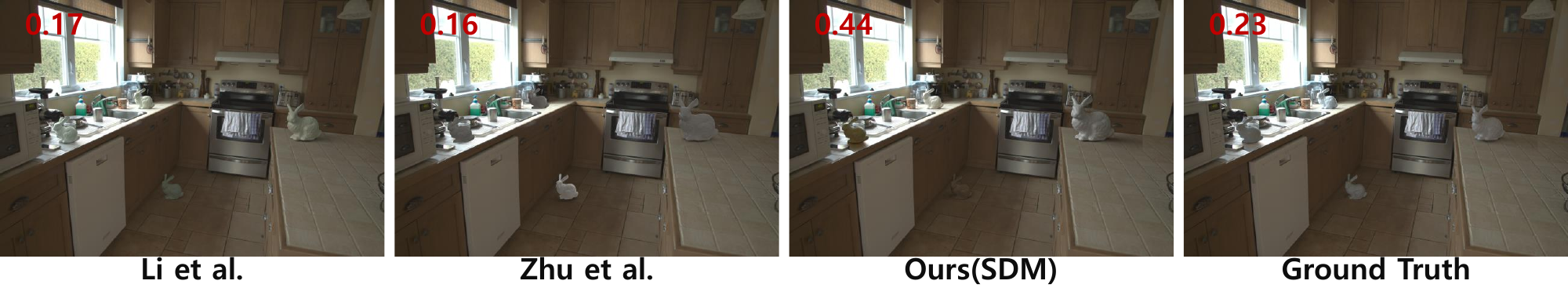}
   \caption{Additional object inversion results on spatially-varying lighting dataset.~\cite{garon19}.}
   \label{fig:supp_oi3}
\end{figure*}

Figs.~\ref{fig:supp_oi1}, \ref{fig:supp_oi2}, and \ref{fig:supp_oi3} present the object insertion results for all 20 images from the spatially-varying lighting dataset~\cite{garon19}. In the user study, participants were asked to evaluate four object insertion images and respond to the question: "Of the four images, please select the one that shows white rabbits blending naturally into the surrounding light." The numbers in the top-left corner of each image indicate the selection rate from the user study, which was conducted with 194 participants. In most cases, our method was preferred by more users compared to baselines. Interestingly, for some images, our results were even preferred over the ground truth. We attribute this to potential discrepancies between the lighting visible in the limited field-of-view RGB image and the actual lighting in the real scene.

\section{Additional Experimental Results}\label{sec:supp:experiments}  
We have included additional experimental results for each dataset.

\subsection{Synthetic Scenes}  
\begin{figure*}[t]
  \centering
  \includegraphics[width=1.0\linewidth]{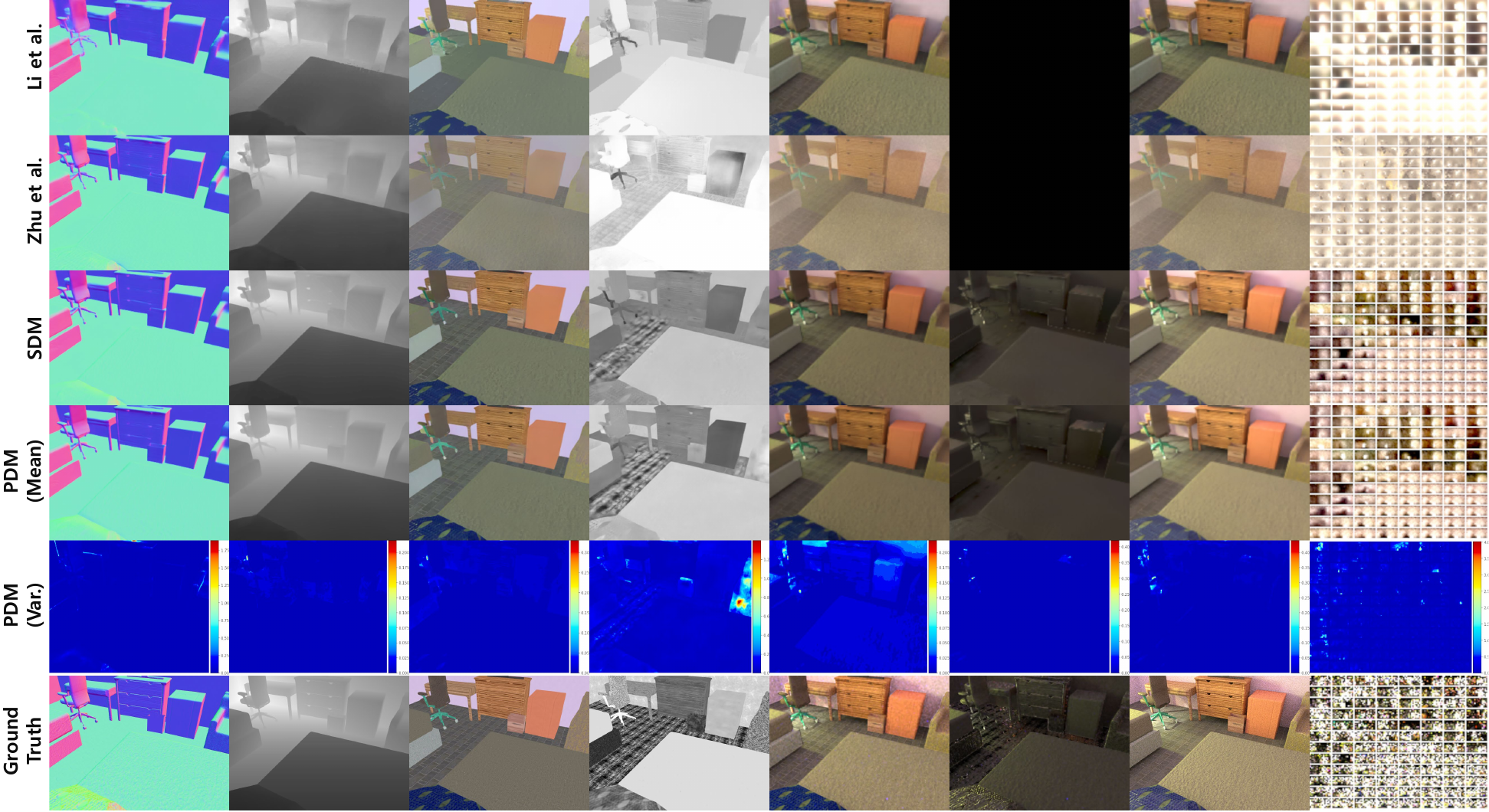}
  \includegraphics[width=1.0\linewidth]{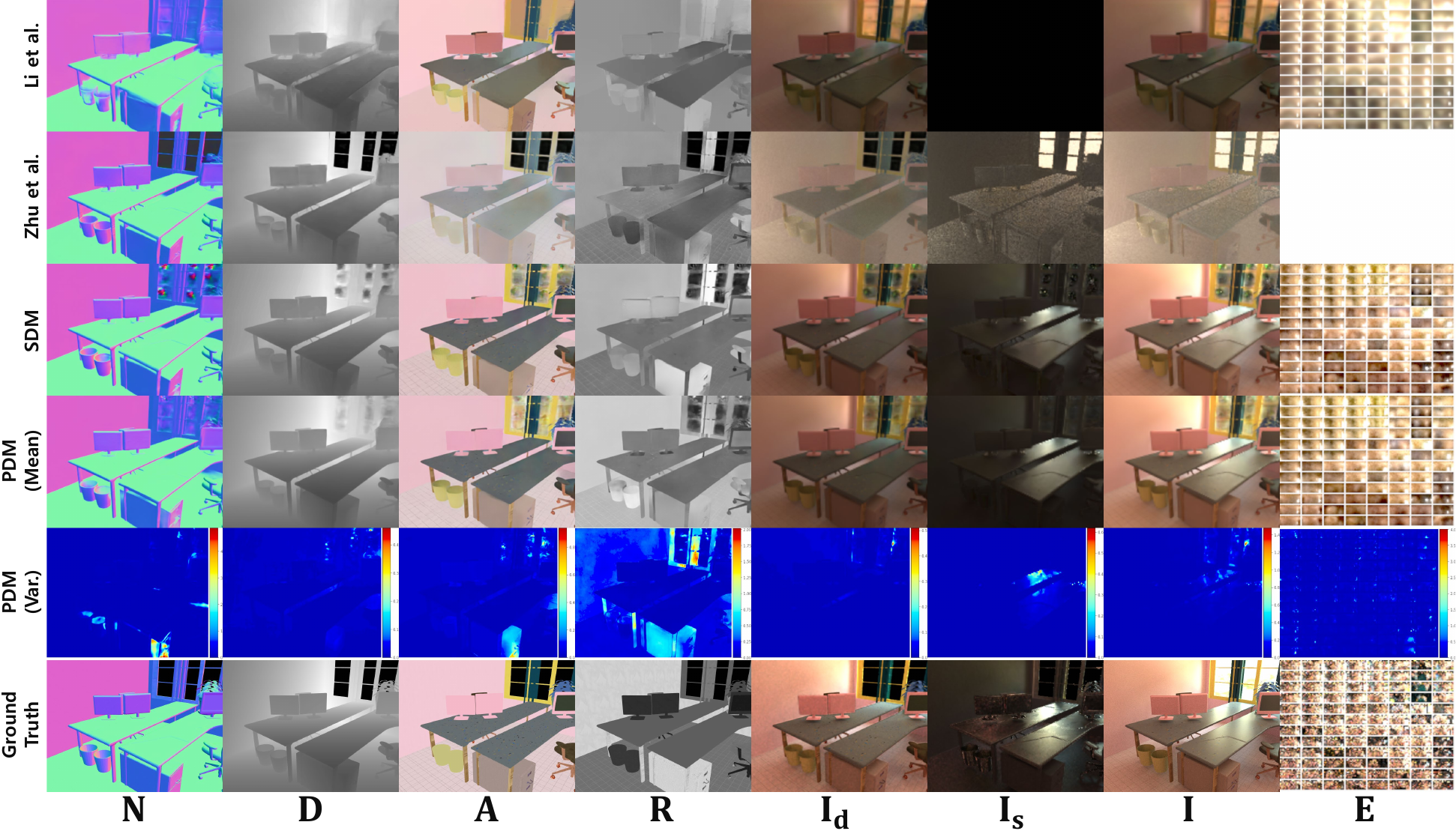}
   \caption{Additional inverse rendering results on OpenRooms FF dataset~\cite{choi2023mair}. }
   \label{fig:supp_ir_indoor}
\end{figure*}

Fig.~\ref{fig:supp_ir_indoor} presents additional experimental results in the synthetic dataset. In the upper sample, strong shadows are cast on the floor, which the baselines fail to remove. In contrast, our SDM and PDM successfully eliminate these shadows. Similarly, for the table in the lower sample, which exhibits strong specular radiance, our method accurately extracts the diffuse albedo.

\subsection{Real-World Scenes}  
\begin{figure*}[t]
  \centering
  \includegraphics[width=1.0\linewidth]{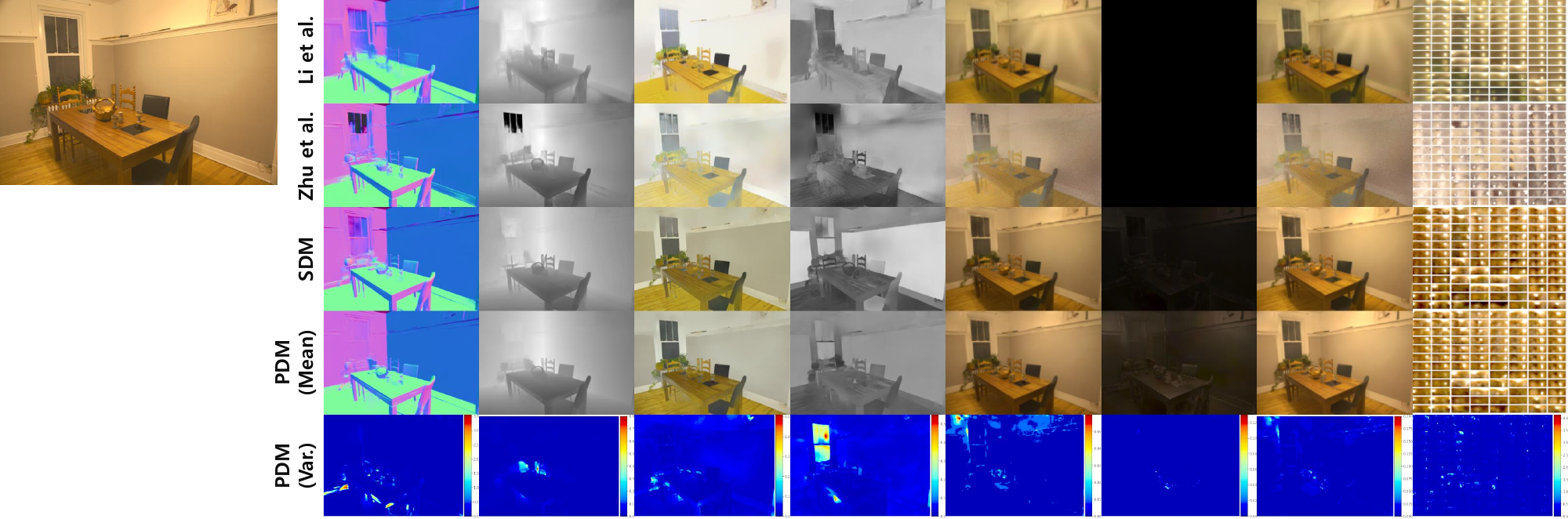}
  \includegraphics[width=1.0\linewidth]{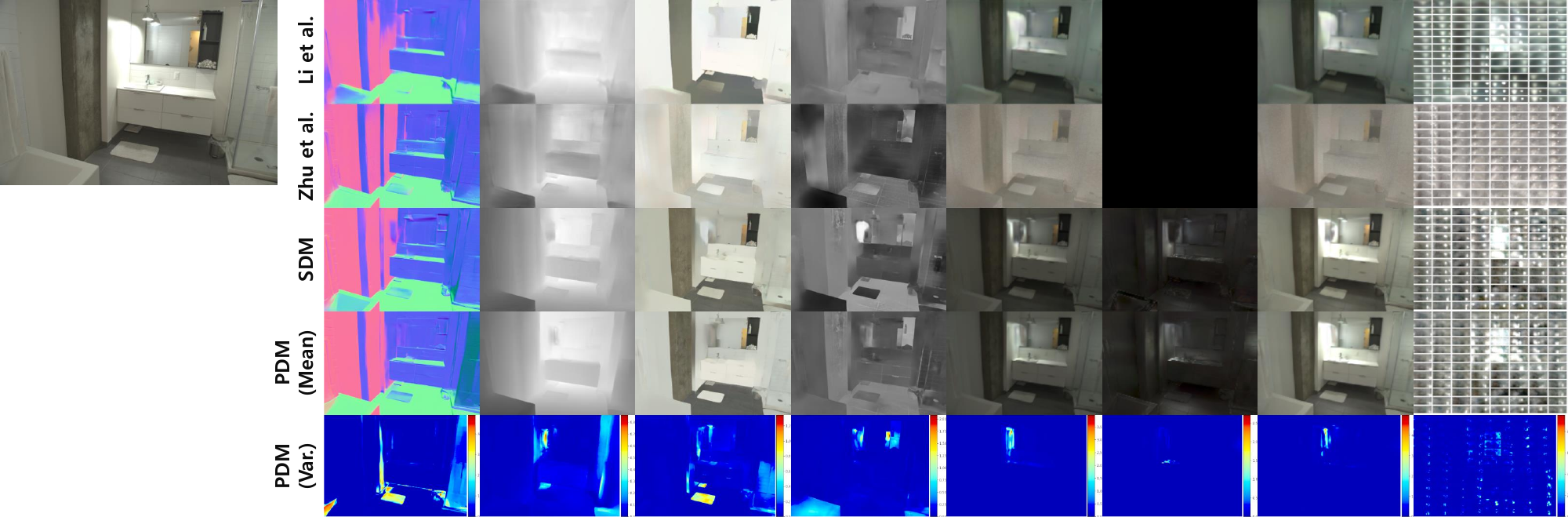}
  \includegraphics[width=1.0\linewidth]{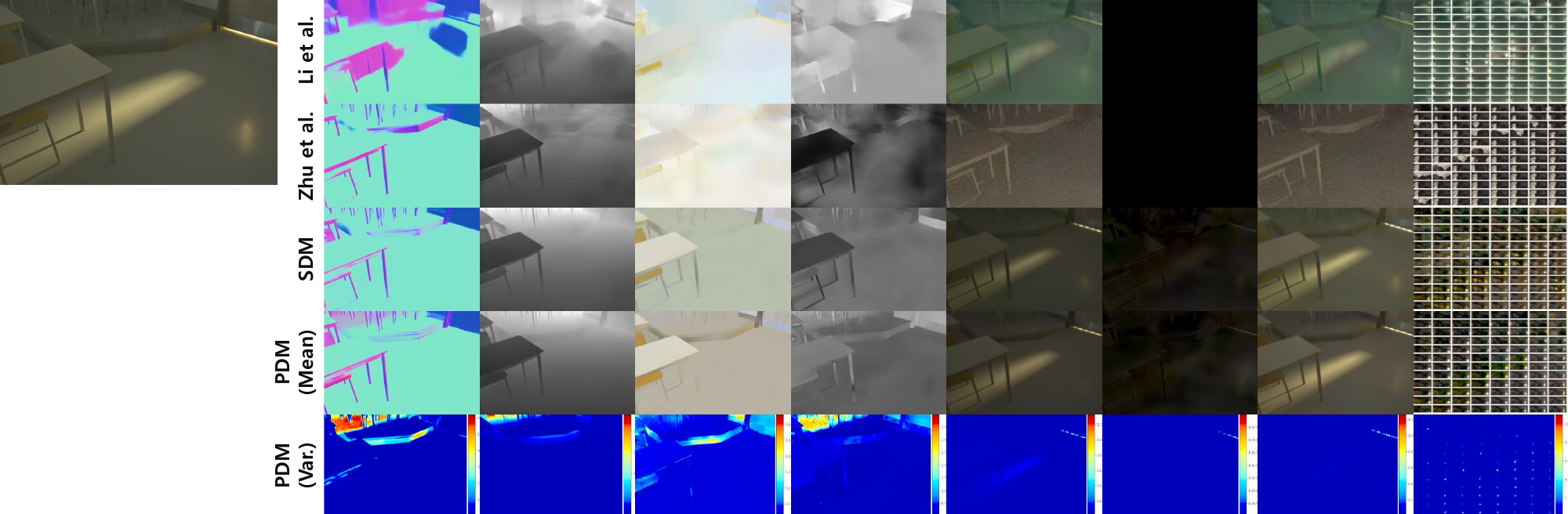}
  \includegraphics[width=1.0\linewidth]{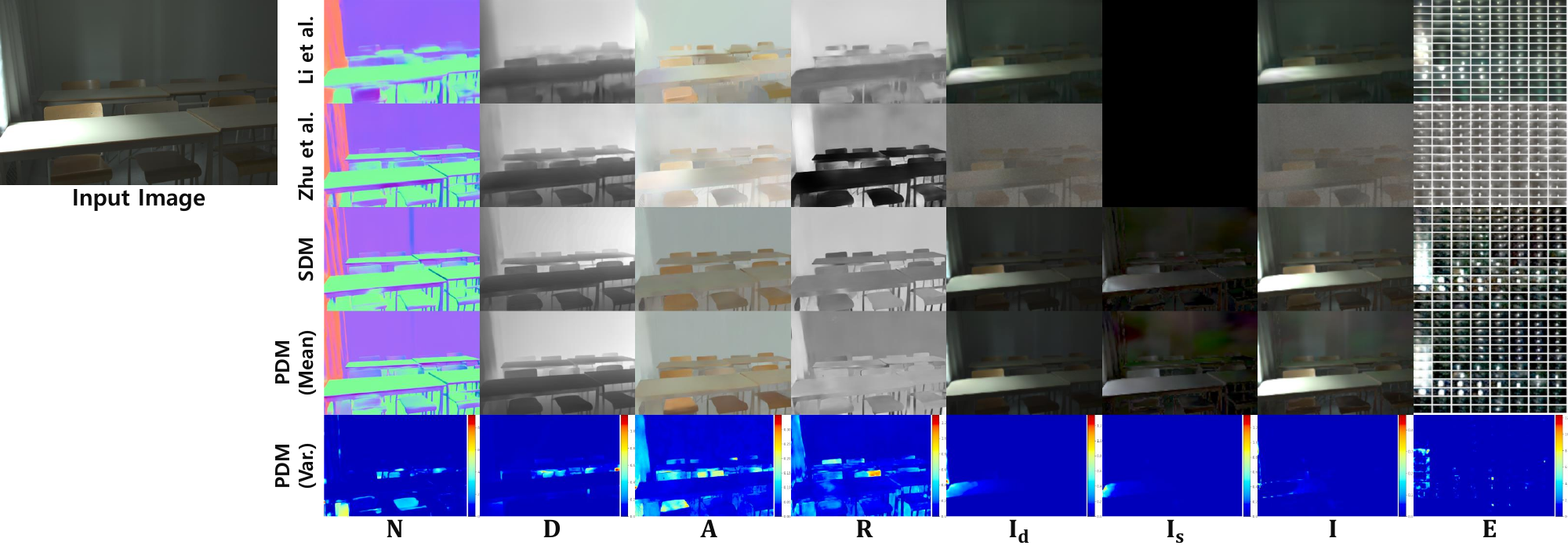}
   \caption{Additional inverse rendering results on spatially-varying lighting dataset.~\cite{garon19}.}
   \label{fig:supp_ir_real}
\end{figure*}

Fig.~\ref{fig:supp_ir_real} shows the inverse rendering results on images from the real-world spatially-varying lighting dataset. In the first sample, \li\ overestimated the albedo of the chair and wall, making them appear too bright, while \zhu\ introduced blotchy artifacts. In contrast, our method accurately predicted the dark regions of the wallpaper and the black albedo of the chair. In the second sample, our method cleanly predicted the albedo of the white mat on the floor. \li\ incorrectly predicted it as black and \zhu displayed artifacts in the albedo of the white wall. In the third sample, \li\ failed to predict the geometry of the floor and \zhu introduced artifacts, while our method consistently predicted both the geometry and albedo. For the fourth sample with strong directional lighting on the white desk, our method predicted a spatially consistent albedo, whereas the baselines exhibited artifacts.

\subsection{Intrinsic Decomposition}  
\begin{figure*}[t]
  \centering
  \includegraphics[width=1.0\linewidth]{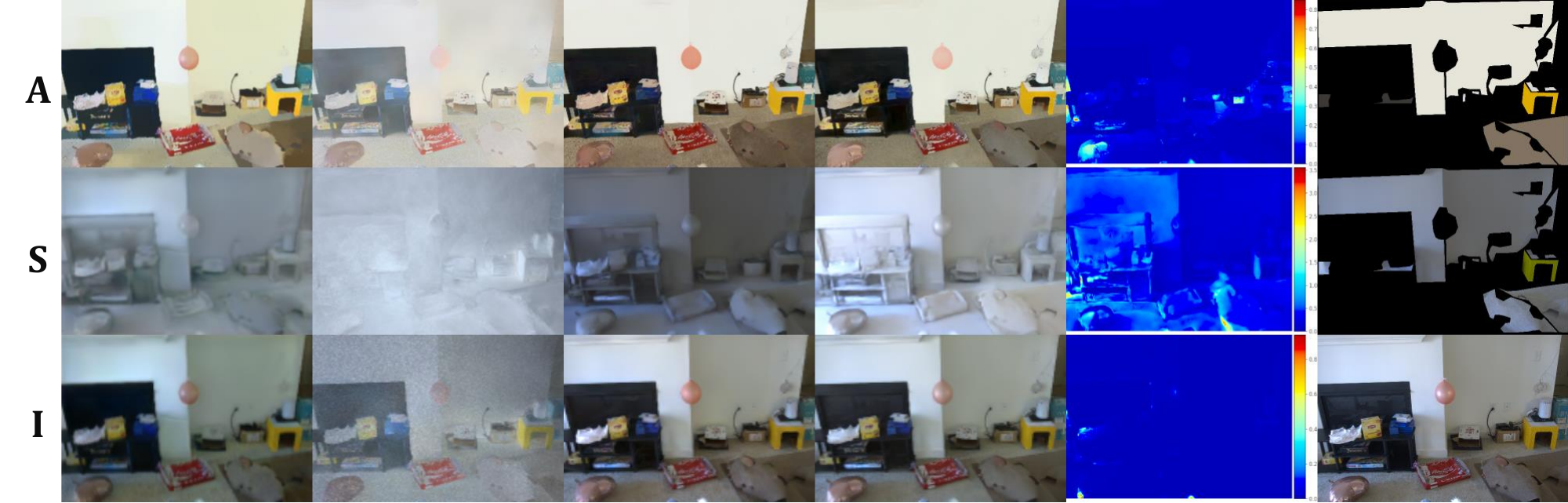}
  \includegraphics[width=1.0\linewidth]{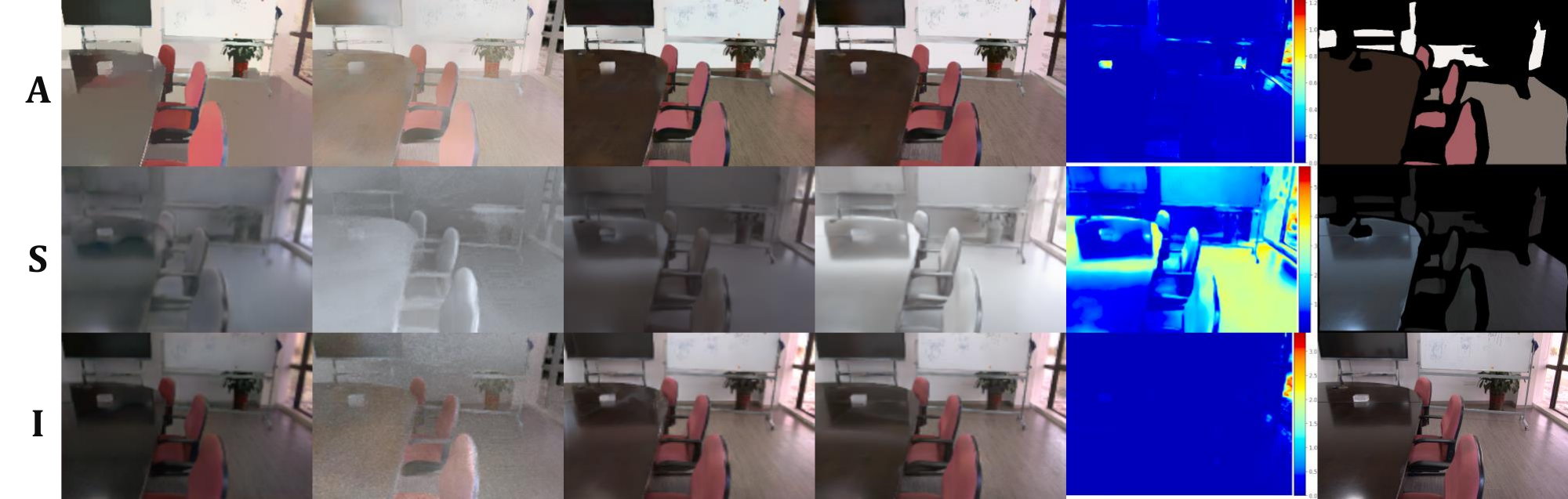}
  \includegraphics[width=1.0\linewidth]{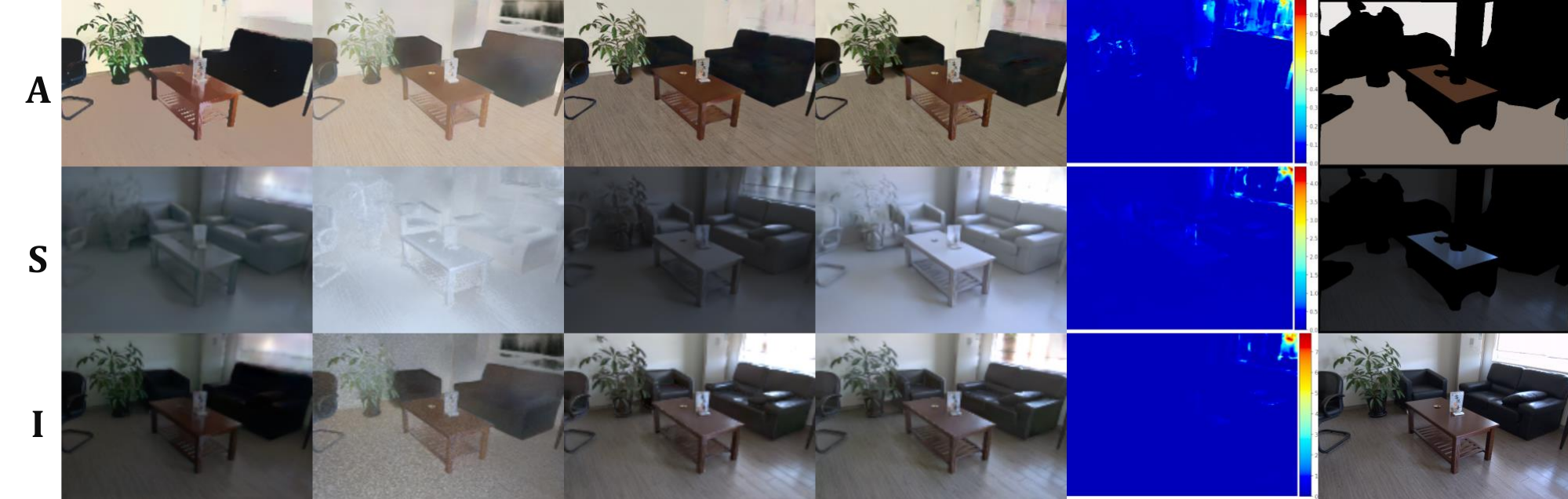}
  \includegraphics[width=1.0\linewidth]{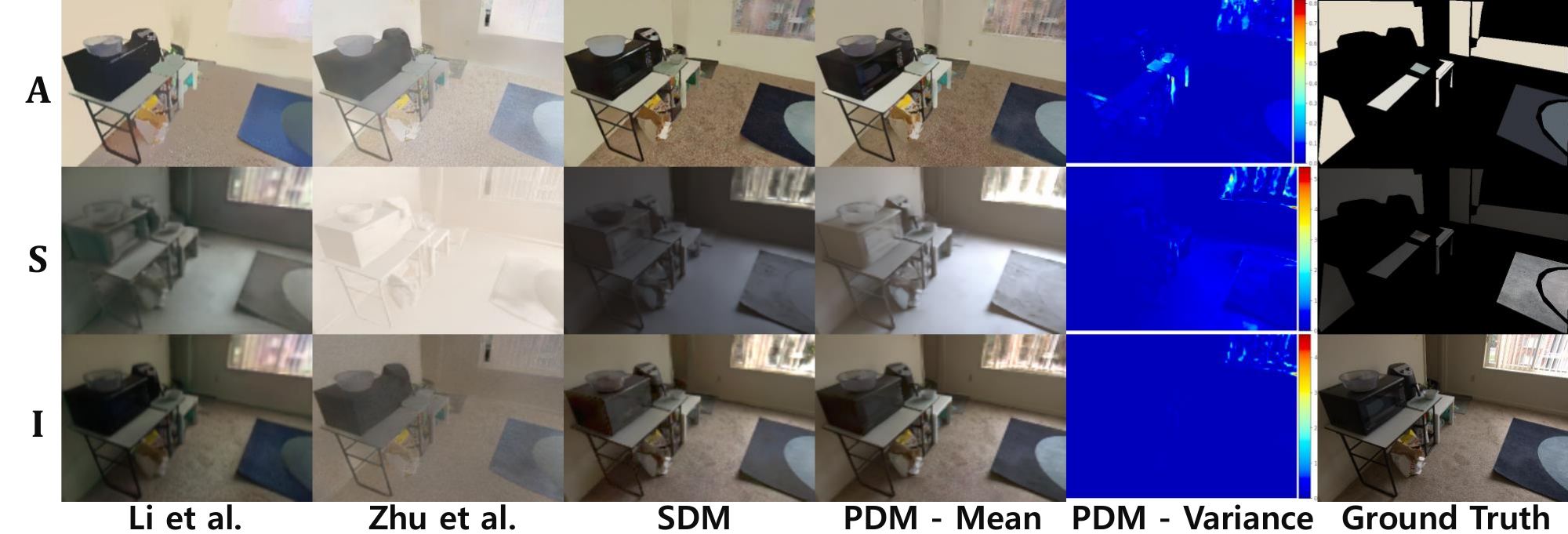}
   \caption{Additional intrinsic decomposition results on MAW dataset~\cite{wu2023maw}.}
   \label{fig:supp_ir_maw}
\end{figure*}

Fig.~\ref{fig:supp_ir_maw} includes additional experimental results on intrinsic decomposition. For the brown carpet in the bottom-right corner of the first sample, unlike the baselines, which predicted an overly bright albedo, our method accurately predicted it. In the second sample, the baselines failed to remove the specular radiance from the brown table, whereas our method effectively removed it. In the third sample, \li\ did not remove the specular radiance from the central table and \zhu showed artifacts in the albedo of the black sofa. In the fourth sample, the baselines struggled to remove shadows from the wall, while our method predicted a spatially consistent albedo.

\subsection{Geometry Prediction}  
\begin{figure*}[t]
  \centering
  \includegraphics[width=0.92\linewidth]{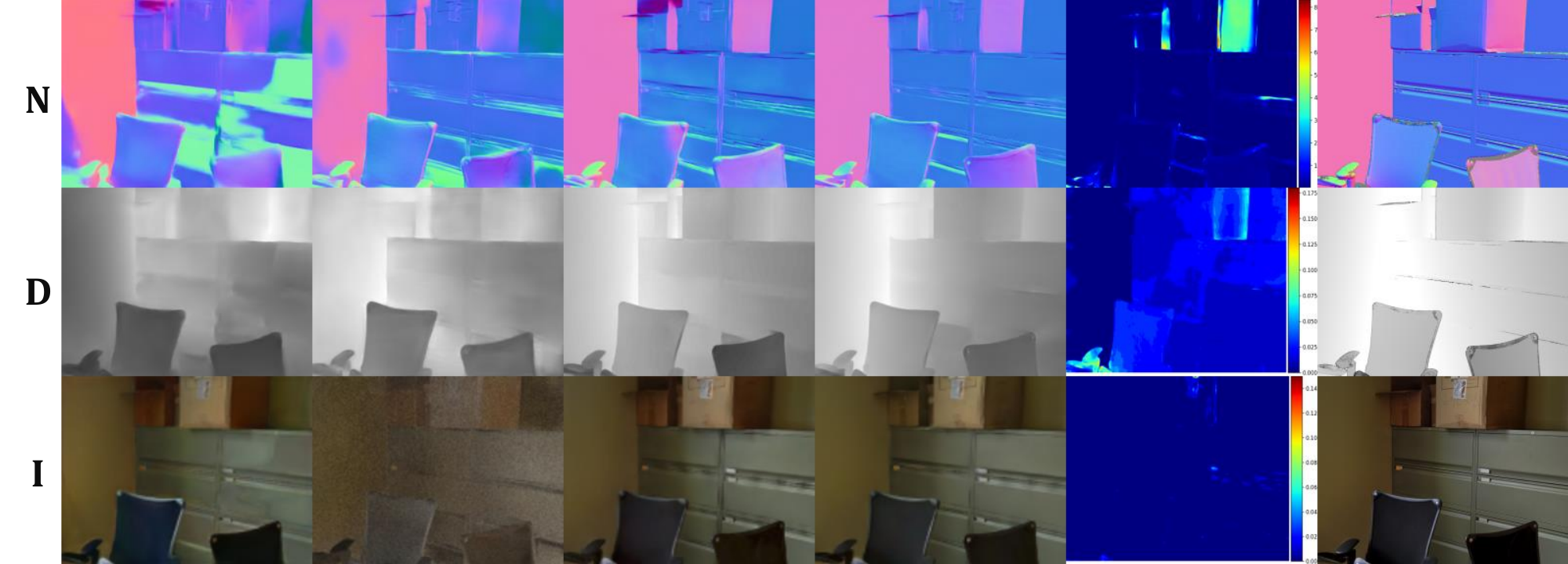}
  \includegraphics[width=0.92\linewidth]{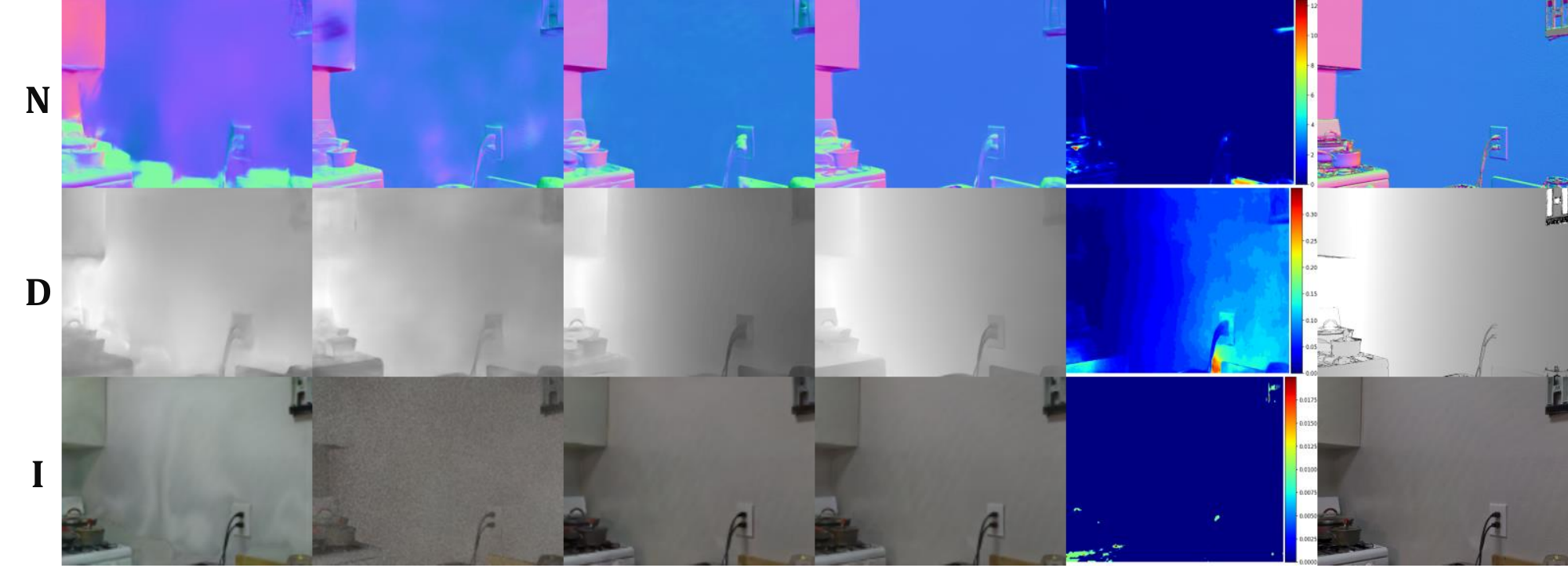}
  \includegraphics[width=0.92\linewidth]{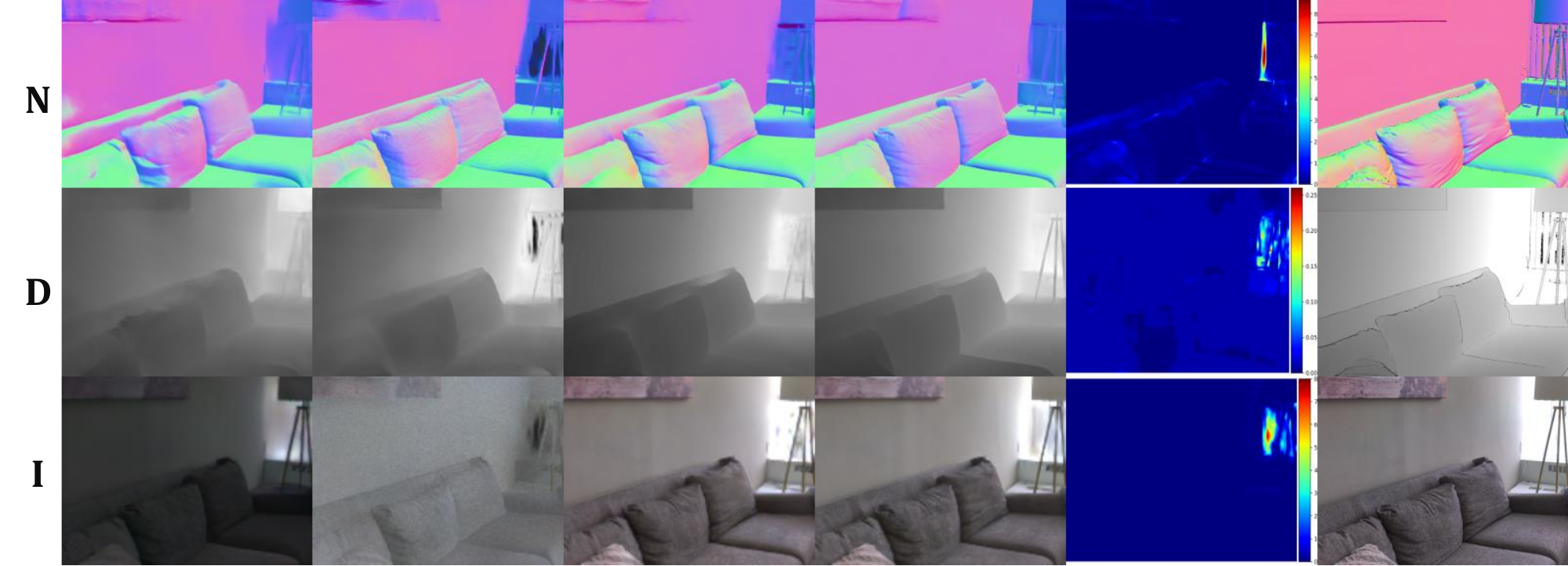}
  \includegraphics[width=0.92\linewidth]{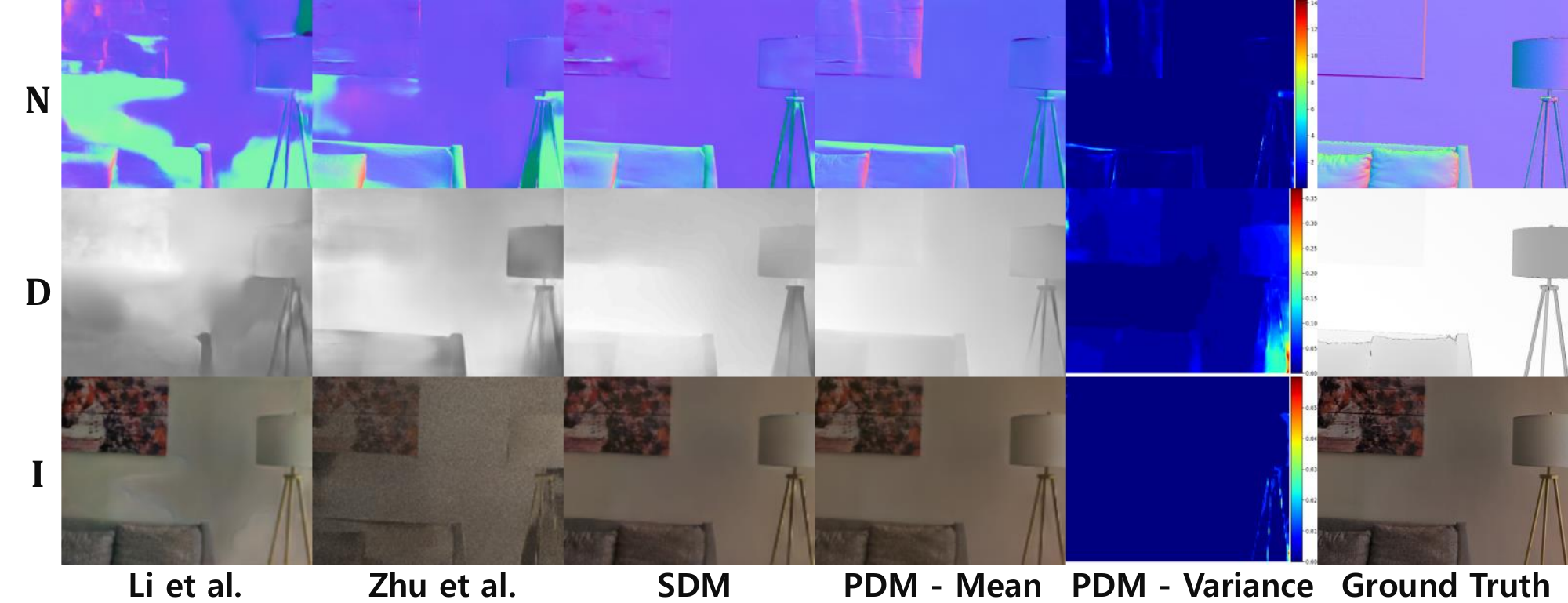}
   \caption{Additional geometry prediction results on DIODE dataset~\cite{vasiljevic2019diode}.}
   \label{fig:supp_ir_diode}
\end{figure*}

Fig.~\ref{fig:supp_ir_diode} presents additional experimental results on geometry prediction. For the cabinet in the first sample, unlike the baselines, which exhibited significant artifacts, our method predicted the geometry relatively accurately. In the second sample, the baselines struggled to understand the context and produced inconsistent geometry predictions. In the third sample, our method demonstrated the fewest artifacts. In the fourth sample, while the baselines displayed numerous artifacts in the normal map predictions for the wall, our method provided comparatively accurate results.

\subsection{Sample Diversity}  
\begin{figure*}[t]
  \centering
  \includegraphics[width=1.0\linewidth]{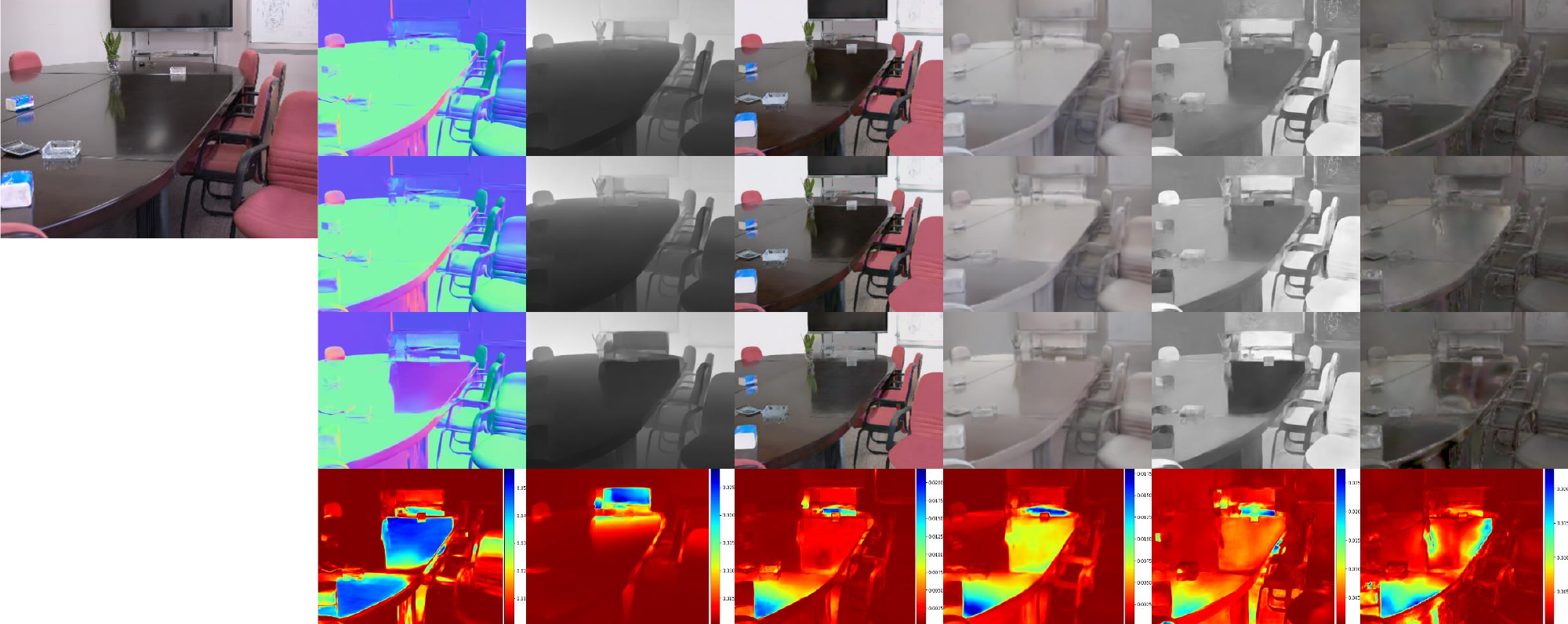}
  \includegraphics[width=1.0\linewidth]{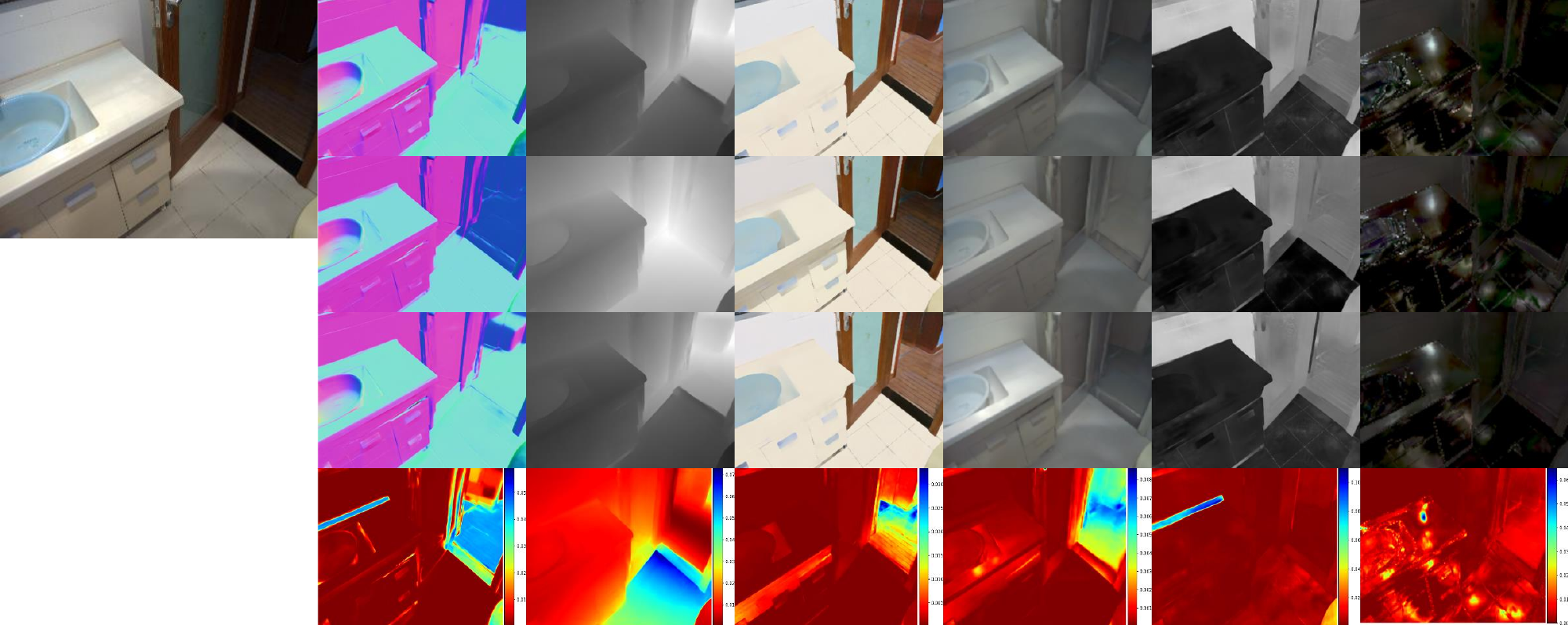}
  \includegraphics[width=1.0\linewidth]{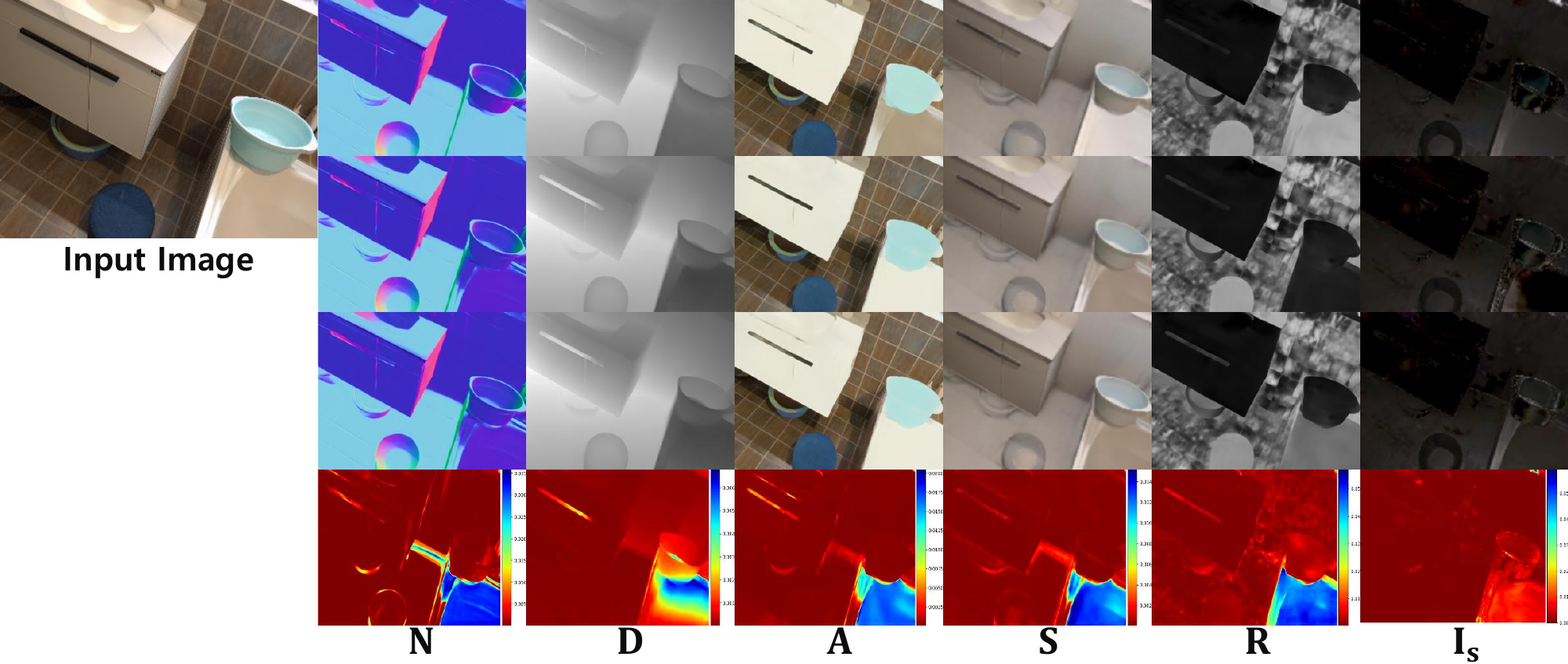}
\caption{Additional experiments on the diversity of samples generated by PDM.}
   \label{fig:supp_diversity_1}
\end{figure*}

\begin{figure*}[t]
  \centering
  \includegraphics[width=1.0\linewidth]{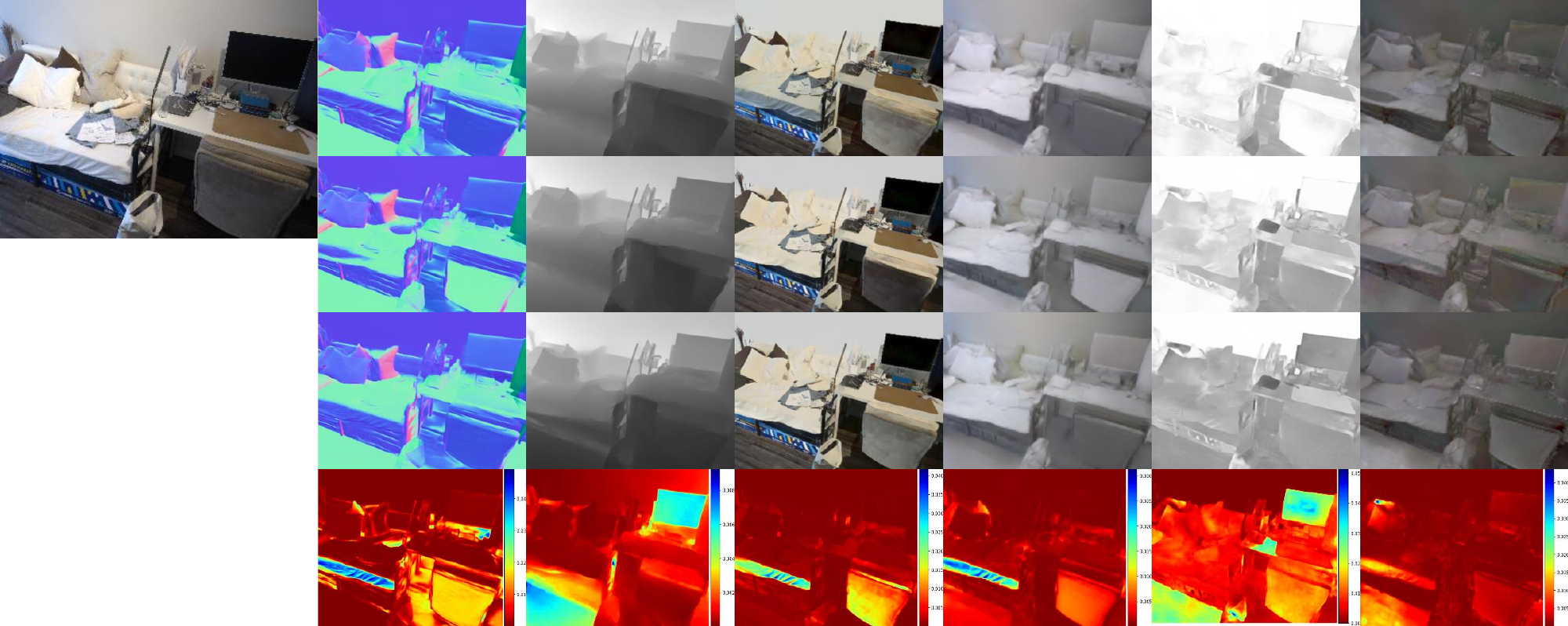}
  \includegraphics[width=1.0\linewidth]{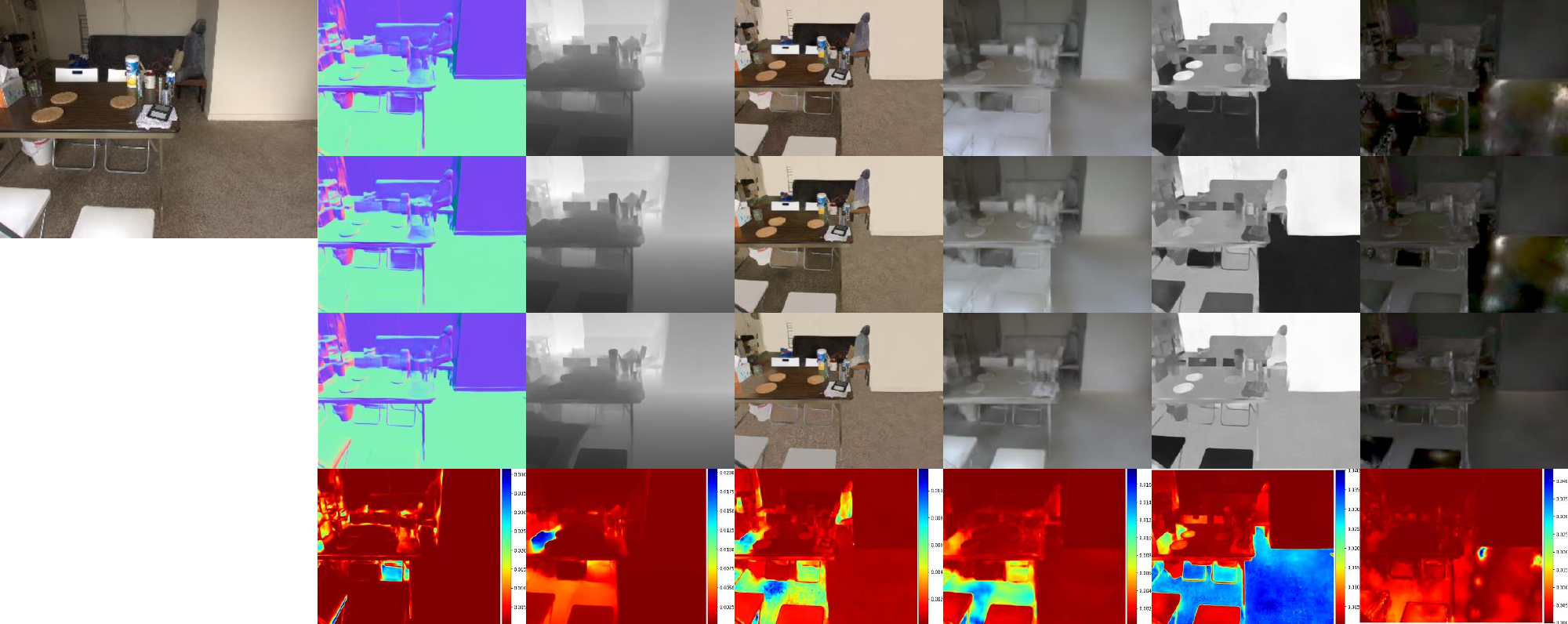}
  \includegraphics[width=1.0\linewidth]{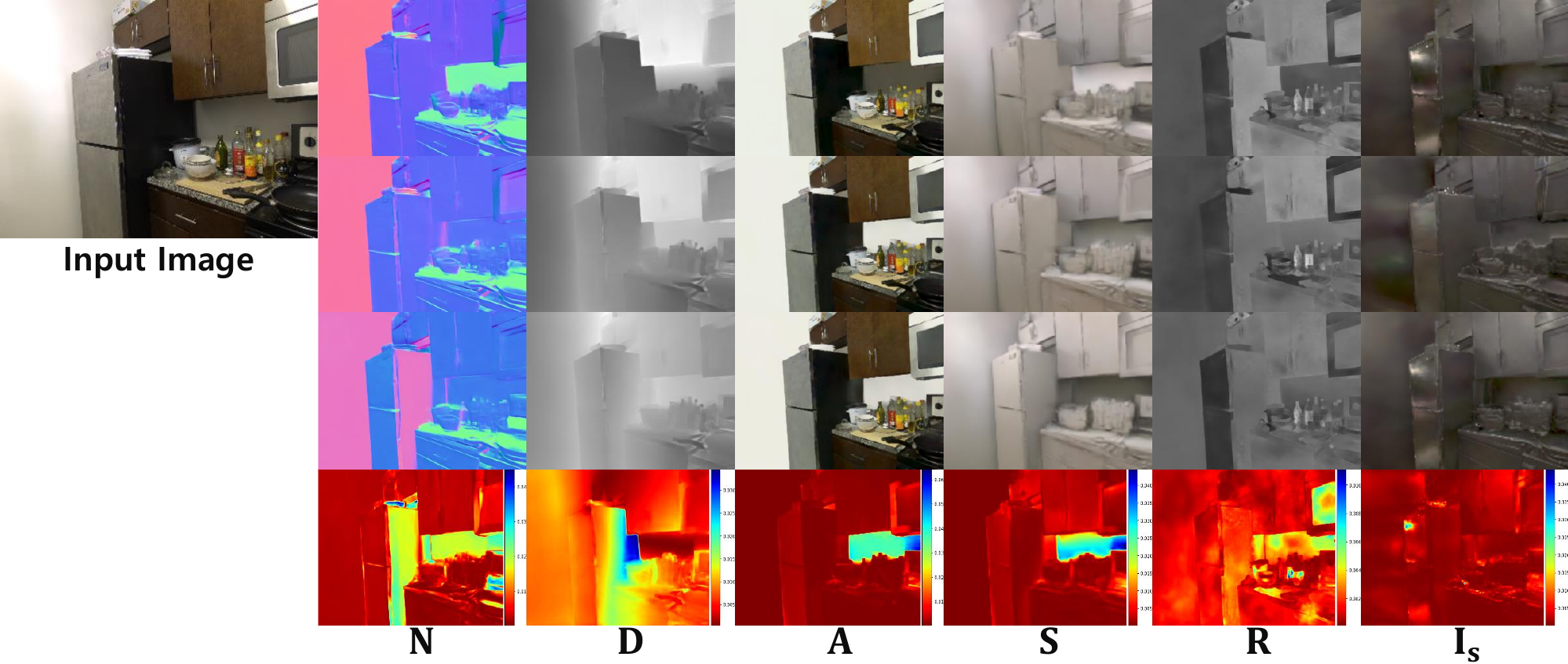}
\caption{Additional experiments on the diversity of samples generated by PDM.}
   \label{fig:supp_diversity_2}
\end{figure*}

Figs.~\ref{fig:supp_diversity_1} and \ref{fig:supp_diversity_2} present experiments that evaluate how effectively PDM can provide a diverse set of possible solutions for inverse rendering. Through the diverse samples generated, we also confirmed that PDM accounts for the dependencies between multiple modalities.

